\documentclass{article}
\usepackage{ijcai24}

\usepackage{times}
\usepackage{soul}
\usepackage{url}
\usepackage[hidelinks]{hyperref}
\usepackage[utf8]{inputenc}
\usepackage[small]{caption}
\usepackage{graphicx}
\usepackage{amsmath}
\usepackage{amsthm}
\usepackage{booktabs}
\usepackage{algorithm}
\usepackage{algorithmic}
\usepackage[switch]{lineno}

\usepackage{mathtools}
\usepackage{mathrsfs}
\usepackage{bm}
\usepackage{microtype}
\usepackage{subfigure}
\usepackage{makecell}
\usepackage{csquotes}
\usepackage{amsfonts}

\title{Self-adaptive PSRO: Towards an Automatic Population-based Game Solver}

\author{
Pengdeng Li$^1$
\and
Shuxin Li$^1$\thanks{Equal contribution.}\and
Chang Yang$^{2}$\footnotemark[1]\and
Xinrun Wang$^{1}$\thanks{Corresponding author.}\and\\
Xiao Huang$^2$\and
Hau Chan$^3$\And
Bo An$^1$\\
\affiliations
$^1$Nanyang Technological University, Singapore\\
$^2$The Hong Kong Polytechnic University, Hong Kong SAR, China\\
$^3$University of Nebraska-Lincoln, Lincoln, Nebraska, United States\\
\emails
\{pengdeng.li, shuxin.li, xinrun.wang, boan\}@ntu.edu.sg, chang.yang@connect.polyu.hk\\xiaohuang@comp.polyu.edu.hk, hchan3@unl.edu
}

\begin{document}

\maketitle

\begin{abstract}
Policy-Space Response Oracles (PSRO) as a general algorithmic framework has achieved state-of-the-art performance in learning equilibrium policies of two-player zero-sum games. However, the hand-crafted hyperparameter value selection in most of the existing works requires extensive domain knowledge, forming the main barrier to applying PSRO to different games. In this work, we make the first attempt to investigate the possibility of self-adaptively determining the optimal hyperparameter values in the PSRO framework. Our contributions are three-fold: (1) Using several hyperparameters, we propose a parametric PSRO that unifies the gradient descent ascent (GDA) and different PSRO variants. (2) We propose the self-adaptive PSRO (SPSRO) by casting the hyperparameter value selection of the parametric PSRO as a hyperparameter optimization (HPO) problem where our objective is to learn an HPO policy that can self-adaptively determine the optimal hyperparameter values during the running of the parametric PSRO. (3) To overcome the poor performance of online HPO methods, we propose a novel offline HPO approach to optimize the HPO policy based on the Transformer architecture. Experiments on various two-player zero-sum games demonstrate the superiority of SPSRO over different baselines.
\end{abstract}

\section{Introduction}

Policy-Space Response Oracles (PSRO)~\cite{lanctot2017unified} since proposed has been the mainstream algorithmic framework for solving two-player zero-sum games. At each epoch, PSRO constructs a meta-game by simulating outcomes of all match-ups of policies of all players and computes the meta-strategies for all players via a meta-solver. It then trains new policies for each player against the opponent's meta-strategy through an oracle and appends the new policies to the player's policy space. The two components -- meta-solver and oracle -- determine the nature of PSRO and various PSRO variants have been proposed~\cite{balduzzi2019open,muller2020generalized,marris2021multi}. Despite the advancements, determining the hyperparameter values in PSRO is non-trivial~\cite{smith2021iterative} and typically involves extensive domain knowledge, which impedes it from broader real-world applications.

Precisely, one needs to determine the meta-solver and the best response (BR) oracle when instantiating PSRO. On one hand, existing works have suggested various meta-solvers such as uniform~\cite{heinrich2016deep}, Nash equilibrium~\cite{lanctot2017unified}, $\alpha$-Rank~\cite{muller2020generalized}, and correlated equilibrium~\cite{marris2021multi}. However, we observe that none of the meta-solvers can consistently beat all the others in terms of learning performance during game solving. On the other hand, the BR policies of a player are typically obtained via a deep reinforcement learning (RL) oracle, e.g., DQN~\cite{mnih2015human}, which involves the initialization and the number of updates of the BR policies. Unfortunately, the determination of these hyperparameter values in most of the existing works is often domain-specific (e.g., poker, soccer). Therefore, an important question is: \textit{Can we automatically determine the optimal hyperparameter values in PSRO?}

In this work, we make the first attempt to answer this question. Specifically, we first propose a parametric PSRO (PPSRO) by introducing two types of hyperparameters: i) \textit{game-free} hyperparameters are the weights of different meta-solvers considered during game solving, and ii) \textit{game-based} hyperparameters are the initialization and the number of updates of a player's BR policies. PPSRO provides a general framework to unify the gradient descent ascent (GDA)~\cite{fiez2021local} and various PSRO variants~\cite{ho1998iterated,balduzzi2019open,muller2020generalized,marris2021multi}. Then, a natural problem is how to determine the hyperparameter values of PPSRO. To solve this problem, we propose a novel framework, self-adaptive PSRO (SPSRO), by casting the hyperparameter value selection of PPSRO as a hyperparameter optimization (HPO) problem where our objective is to learn an HPO policy that can self-adaptively select the optimal hyperparameter values of PPSRO during game solving. A straightforward method to optimize the HPO policy is to use online approaches such as Optuna~\cite{akiba2019optuna}. Unfortunately, online HPO methods only use online generated data (past epochs of SPSRO), typically constraining the training objectives to be cheaply computable~\cite{cheny2022towards} and the performance could be poor. To overcome these limitations, we propose an offline HPO approach to optimize the HPO policy based on the Transformer architecture~\cite{vaswani2017attention,chen2021decision}. Specifically, we formulate the HPO policy optimization as a sequence modeling problem where a Transformer model is trained by using an offline dataset and then used to predict the hyperparameter values conditioned on past epochs of SPSRO. Intuitively, a well-trained HPO policy has the potential to transfer to different games, reducing the effort needed for researchers to conduct the costly hyperparameter tuning when applying PSRO to various games.

In summary, the contributions of this work are three-fold: (1) By introducing several hyperparameters, we propose a parametric version of PSRO (PPSRO) which unifies GDA and various PSRO variants. (2) We propose a novel framework, self-adaptive PSRO (SPSRO), by formulating an optimization problem where our objective is to learn an HPO policy that can self-adaptively determine the optimal hyperparameter values of PPSRO. (3) To overcome the poor performance of classic online HPO methods, we propose an offline HPO approach to optimize the HPO policy based on the Transformer architecture. We evaluate the effectiveness of our approach through extensive experiments on a set of two-player zero-sum normal-form and extensive-form games, and the results demonstrate that SPSRO with Transformer can achieve significantly better learning performance than different baselines.

\section{Related Works}

PSRO~\cite{lanctot2017unified} generalizes the double oracle algorithm~\cite{mcmahan2003planning} and unifies various multi-agent learning methods including the fictitious play (FP)~\cite{robinson1951iterative,brown1951iterative}, neural fictitious self-play (NFSP)~\cite{heinrich2016deep} (an extension of FP in the context of deep RL), iterated best response (IBR)~\cite{ho1998iterated}, and independent reinforcement learning (InRL)~\cite{matignon2012independent}. Recently, many works have been done toward improving PSRO, including the scalability~\cite{mcaleer2020pipeline,smith2021iterative}, diversity of BRs~\cite{perez2021modelling,liu2021towards}, the introduction of novel meta-solvers, e.g., $\alpha$-Rank~\cite{muller2020generalized}, correlated equilibrium~\cite{marris2021multi}, and neural meta-solver~\cite{feng2021neural}, and application to mean-field games~\cite{muller2022learning}. Moreover, the challenging strategy exploration problem has also been extensively investigated~\cite{wellman2006methods,schvartzman2009stronger,schvartzman2009exploring,jordan2010strategy,wang2022evaluating}. Despite the advancements, a critical observation we obtained is that given a set of meta-solvers, none of them can consistently beat (dominate) all the others in terms of learning performance during game solving (in the sense that we just evaluate PSRO as an online algorithm). On the other hand, the BRs of a player are typically obtained via a deep RL oracle such as DQN~\cite{mnih2015human}, where the hyperparameters (e.g., the initialization and the number of updates) are often domain-specific (e.g., poker, soccer) and most of the existing works manually select the hyperparameter values based on domain knowledge. In this work, we develop a novel framework to self-adaptively determine the optimal hyperparameter values in PSRO, which can be transferred to different games without fine-tuning.

Another line of related work is hyperparameter optimization (HPO). Existing works on HPO can be roughly categorized into online and offline HPO. The classic online HPO methods include Bayesian optimization~\cite{snoek2012practical} and its variants~\cite{krause2011contextual,bardenet2013collaborative,swersky2013multi,feurer2015initializing,volpp2019meta,wistuba2020few,rothfuss2021pacoh}, and recurrent neural networks (RNNs)~\cite{duan2016rl,wang2016learning,chen2017learning}. However, online HPO methods only use online generated data, typically constraining the training objectives to be cheaply computable and the performance could be poor. Our work is closely related to~\cite{cheny2022towards} which proposes the first offline Transformer-based HPO method. Nevertheless, it is non-trivial to optimize the HPO policy as it involves several critical challenges such as how to generate an offline dataset for training. To our knowledge, this work is the first attempt to explore and develop a self-adaptive hyperparameter value selector in game theory.

\section{Preliminaries}

In this section, we first present the game definition and then the procedure of the PSRO algorithm.

\subsection{Games}

Consider a two-player zero-sum game represented by a tuple $G = (\mathcal{N}, \mathcal{S}, \mathcal{A}, p, \{r^i\}_{i \in \mathcal{N}}, \mathcal{T})$, where players are indexed by $\mathcal{N}=\{1, 2\}$. Let $N=\vert\mathcal{N}\vert=2$. $\mathcal{S}$ and $\mathcal{A}$ denote the players' state and action spaces, respectively. $\mathcal{T} = \{0, 1, \cdots, T\}$ is time index set. At $t \in \mathcal{T}$, player $i$ in state $s_t^i \in \mathcal{S}$ takes an action $a_t^i \in \mathcal{A}$ and then changes to new state $s_{t+1}^i \sim p(\cdot \vert \bm{s}_t, \bm{a}_t)$ and receives a reward $r^i(\bm{s}_t, \bm{a}_t)$, where $\bm{s}_t = (s_t^i)_{i \in \mathcal{N}}$ and $\bm{a}_t = (a_t^i)_{i \in \mathcal{N}}$ are respectively joint state and joint action of all players, $p: \mathcal{S}^N \times \mathcal{A}^N \to \Delta(\mathcal{S})$\footnote{$\Delta(\mathcal{X})$ denotes the probability distribution over the space $\mathcal{X}$.} is the transition function and $r^i: \mathcal{S}^N \times \mathcal{A}^N \to \mathbb{R}$ is the reward function with $\sum_{i \in \mathcal{N}} r^i(\bm{s}_t, \bm{a}_t) = 0$. Let $\pi^i: \mathcal{S} \to \Delta(\mathcal{A})$ denote the player $i$'s policy (strategy)\footnote{We interchangeably use \textit{policy} and \textit{strategy} in this work.} with $\pi^i \in \Pi^i$ where $\Pi^i$ is the policy space. Given the joint policy of all players $\bm{\pi} = (\pi^i)_{i \in \mathcal{N}} \in \Pi = \times_{i \in \mathcal{N}} \Pi^i$, each player $i$ aims to maximize his own value function $V^i(\bm{\pi}, \bm{s}_0)=\mathbb{E} \big[ \sum_{t=0}^{T} r^i(\bm{s}_t, \bm{a}_t)\vert \bm{a}_t\sim\bm{\pi}, \bm{s}_{t+1}\sim p\big]$, where $\bm{s}_0$ is the players' initial states. 

A mixed strategy $\sigma^i \in \Delta(\Pi^i)$\footnote{In principle $\Pi^i$ could be an infinite set. However, $\Pi^i$ is typically iteratively expanded by learning algorithms such as PSRO and hence, is considered finite in this work.} is called a \textit{meta-strategy} which is the probability distribution over the player $i$'s policy space $\Pi^i$. More precisely, suppose that there are $c \ge 1$ policies in player $i$'s policy space, then the meta-strategy of $i$ is $\sigma^i = (\sigma^{i,1}, \cdots, \sigma^{i,c})$ with $\sigma^{i,j} \ge 0$ and $\sum_{j=1}^c \sigma^{i,j} = 1$. Accordingly, $\bm{\sigma} = (\sigma^i)_{i \in \mathcal{N}} \in \Delta(\Pi)$ is the joint meta-strategy of all players. Given the joint meta-strategy of all players except $i$, $\bm{\sigma}^{-i}$, the expected payoff to player $i$'s policy $\pi^i \in \Pi^i$ is given by $R^i(\pi^i, \bm{\sigma}^{-i}) = \sum_{\bm{\pi}^{-i} \in \Pi^{-i}} \bm{\sigma}^{-i}(\bm{\pi}^{-i}) V^i(\pi^i, \bm{\pi}^{-i})$ and the set of \textit{best responses (BRs)} of player $i$ is defined as $\text{BR}^i(\bm{\sigma}^{-i}) = \arg\max_{\pi^i \in \Pi^i} R^i(\pi^i, \bm{\sigma}^{-i})$. The quality of $\bm{\sigma}$ can be measured by the NashConv~\cite{lanctot2017unified}. For player $i \in \mathcal{N}$, the NashConv is defined as $\mathcal{R}^i(\bm{\sigma}) = R^i(\text{BR}^i(\bm{\sigma}^{-i}), \bm{\sigma}^{-i}) - \sum_{\pi^i \in \Pi^i} \sigma^i(\pi^i) R^i(\pi^i, \bm{\sigma}^{-i})$. That is, the NashConv of player $i$ is the gain he can obtain when he unilaterally deviates from the current joint meta-strategy to a (pure) BR strategy. Therefore, the quality of $\bm{\sigma}$ is measured by the total NashConv of all players $\mathcal{R}(\bm{\sigma}) = \sum_{i \in \mathcal{N}}\mathcal{R}^i(\bm{\sigma})$.

\subsection{Policy-Space Response Oracles}

Given a game, PSRO first initializes the policy space of each player $\Pi^i$ using randomly generated policies and then expands the policy space in three iterated phases\footnote{Note that throughout this work, $e$ always represents the index of PSRO epoch, neither the power of $e$ nor the Euler's number.} (as shown in Figure~\ref{fig:psro}): (1) Synthesize a meta-game with all match-ups of policies of all players and compute the missing payoff entries in the payoff tensor $M$ via simulation. (2) Compute the joint meta-strategy $\bm{\sigma}$ using a meta-solver $\mathcal{M}$ on the synthesized meta-game $M$. Different meta-solvers can be used during training, e.g., Nash equilibrium~\cite{lanctot2017unified}, correlated equilibrium~\cite{marris2021multi}, $\alpha$-Rank~\cite{muller2020generalized}, and uniform distribution~\cite{heinrich2016deep}. (3) Compute each player $i$'s BR $\pi^{i, \text{BR}}$ using an oracle $\mathcal{O}^i$ given the joint meta-strategy $\bm{\sigma}$ and add the BR $\pi^{i, \text{BR}}$ to player $i$'s policy space $\Pi^i$. As the other player's policy spaces $\Pi^{-i}$ and joint meta-strategy $\bm{\sigma}^{-i}$ are fixed, the computation of BR is a single-player optimization problem from player $i$'s perspective. In practice, the best response is typically approximated by using deep RL algorithms, e.g., DQN~\cite{mnih2015human}. Specifically, the best response $\pi^{i, \text{BR}}$ is trained for $K$ updates, given that the other player uses the policy $\bm{\pi}^{-i} \sim \bm{\sigma}^{-i}$, i.e., $\bm{\pi}^{-i}$ is sampled according to $\bm{\sigma}^{-i}$.

\section{Motivating Example\label{sec:motivating-example}}

\begin{figure}[ht]
\centering
\includegraphics[width=\columnwidth]{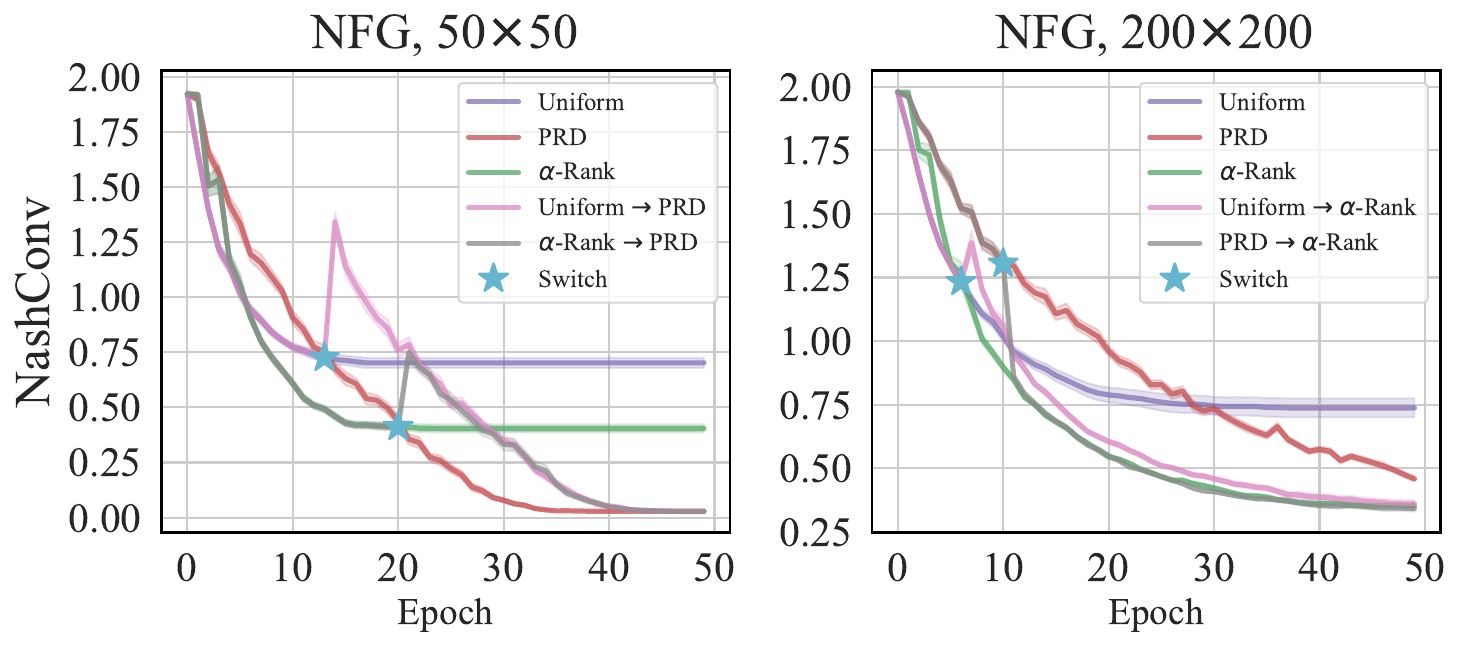}
\caption{NashConv of different PSRO runs.}
\label{fig:motivating-example}
\end{figure}

In this section, we provide some examples to better illustrate the motivation of this work. Consider a two-player zero-sum normal-form game (NFG) of size $\vert \mathcal{A}_1 \vert \times \vert \mathcal{A}_2 \vert$. Let $\bm{\mathcal{M}}$ denote the set of meta-solvers of interest. In this example (as well as this work), we consider the most commonly used three meta-solvers: Uniform~\cite{heinrich2016deep}, $\alpha$-Rank~\cite{muller2020generalized}, and PRD~\cite{lanctot2017unified}. We conduct two types of experiments: i) consistently using a single meta-solver during the PSRO procedure, and ii) switching the meta-solver from one to another at some intermediate iteration of the PSRO. The results are shown in Figure~\ref{fig:motivating-example}.

From the results, we can obtain the following observations. (1) For the three meta-solvers, none of them can consistently beat (dominate) all the others in terms of NashConv during the PSRO procedure. For instance, at the early stage of the PSRO procedure, Uniform performs better than the other two meta-solvers in terms of NashConv. However, it only converges to a high NashConv value, which is also observed in previous works~\cite{muller2020generalized}. (2) By switching from one meta-solver to another during the PSRO procedure, we can achieve better learning performance in terms of NashConv. Moreover, the comparison between the two cases: \enquote{$\alpha$-Rank $\to$ PRD} and \enquote{PRD $\to$ $\alpha$-Rank}, again verifies the previously observed fact that none of the meta-solvers can consistently beat all the others in terms of NashConv during the PSRO procedure\footnote{These observations are different from~\cite{wang2022evaluating} which regards the strategy exploration and its evaluation as two orthogonal components. For example, one can use PRD to guide the BR policies computation but use $\alpha$-Rank to compute the meta-distribution for decision-making after the new BR policies of players are added to their respective policy spaces. However, it is worth noting that our observations do not cause inconsistency with~\cite{wang2022evaluating} as we just evaluate PSRO as an online algorithm.}.

The above observations motivate us to think about a natural question: How to determine the meta-solvers during the PSRO procedure such that we can obtain better learning performance? Note that the examples in Figure~\ref{fig:motivating-example} are NFGs. For extensive-form games (EFGs), in addition to the meta-solver, the hyperparameters also include the initialization of a BR policy $\pi^{i, \text{BR}}$ and the number of updates $K$ for training the BR policy. Most of the existing works determine the values of these hyperparameters by hand-crafted tuning, which typically requires extensive domain knowledge.

Thus, a critical problem to be addressed is: \textit{how to automatically determine the optimal hyperparameter values during the PSRO running?} Specifically, at each epoch, we need to i) choose one or multiple meta-solver(s), ii) determine how to initialize the new BR policies, e.g., random initialization, copy from one of the previous BRs, or mix, and iii) determine the number of updates $K$ of the new BR policy of each player. In this work, we make the first attempt to develop a novel framework that can self-adaptively determine the optimal hyperparameter values during the PSRO running.

\section{Self-adaptive PSRO}

In this section, we establish the Self-adaptive PSRO (SPSRO) through two steps: (1) We parameterize the PSRO algorithm (PPSRO) by introducing several hyperparameters. (2) We cast the hyperparameter value selection of PPSRO as a hyperparameter optimization (HPO) problem where our objective is to learn an HPO policy that will self-adaptively determine the optimal hyperparameter values of PPSRO. 

\subsection{Parametric PSRO}

First, inspired by the observations in the previous section, in this work, instead of considering a single meta-solver as most of the existing works, we use the meta-solver set $\bm{\mathcal{M}}^{\bm{\alpha}}$ with $m$ different meta-solvers and associate it with a vector $\bm{\alpha} = (\alpha_1, \cdots, \alpha_m)$ specifying the weight of each meta-solver. Intuitively, by combining multiple meta-solvers, we could obtain better performance. As $\bm{\alpha}$ is only dependent on the meta-game payoff tensor $M$ regardless of the underlying games (normal-form or extensive-form games), we refer to the weights in it as \textit{game-free} hyperparameters.

Second, we introduce a parametric BR oracle $\mathcal{O}^i(\bm{\sigma}; \beta, K)$ where the hyperparameters include\footnote{There could be other hyperparameters such as batch size in the BR oracle. Nevertheless, as the first attempt to explore the possibility of self-adaptively determining the optimal hyperparameter values, we focus on the ones that enable us to unify various PSRO variants. Moreover, it is worth noting using the same type of BR oracle (DQN in this work) with the same configuration for all the other hyperparameters is important to ensure a fair comparison between our approach and baselines. See Appendix A for more discussion.}: (1) the initialization parameter $\beta \in [0, 1]$ which determines the initialization of the new BR policy of player $i$ by mixing the BR policy obtained in the last epoch with a randomly initialized policy $\pi^{i, \text{random}}$, and (2) the number $K$ which determines the number of updates needed for training the new BR policy. Formally, at each epoch $e$, we initialize player $i$'s BR policy as $\pi^{i, \text{BR}, e} = \beta \pi^{i, \text{BR}, e-1} + (1 - \beta) \pi^{i, \text{random}}$, and then update this BR policy $\pi^{i, \text{BR}, e}$ for $K$ steps. After that, the trained BR policy $\pi^{i, \text{BR}, e}$ is added to player $i$'s policy space $\Pi^i$. As $\beta$ and $K$ are highly dependent on the underlying games (e.g., poker, soccer), we refer to them as \textit{game-based} hyperparameters.

By specifying $\bm{\alpha}$, $\beta$ and $K$, we can obtain GDA and various PSRO variants (Table~\ref{tab:gda_psro}). For example, GDA can be instantiated as follows. Suppose that $\mathcal{M}_b$ is the meta-solver \enquote{Last-One}, then we set $\alpha_b = 1$ and $\alpha_{d \ne b}= 0$, which implies that the meta-strategy of player $i$ is $\sigma_b^i = (\sigma_b^{i, 1} = 0, \cdots, \sigma_b^{i, e-1} = 1)$ at epoch $e$. Then, player $i$ initializes the BR policy $\pi^{i, \text{BR}, e}$ with $\beta = 1$ and trains $\pi^{i, \text{BR}, e}$ with $K=1$ update.

\begin{table}[ht]
\centering
\begin{tabular}{lll}
\toprule
\textbf{Algorithm} & $\bm{\mathcal{M}}^{\bm{\alpha}}$ &$\mathcal{O}^i(\bm{\sigma}; \beta, K)$\\
 \midrule
GDA & Last-One & $\beta = 1$ \& $K=1$\\
InRL & Last-One & $\beta = 1$ \& $K=\bar{K}$\\
PSRO$_{\text{P}}$ & Penultimate & $\beta \in [0, 1]$ \& $K=\bar{K}$\\
PSRO$_{\text{U}}$ & Uniform  & $\beta \in [0, 1]$ \& $K=\bar{K}$ \\
PSRO$_{\text{N}}$ & Nash & $\beta \in [0, 1]$ \& $K=\bar{K}$\\
PSRO$_{\text{rN}}$ & Rectified Nash & $\beta \in [0, 1]$ \& $K=\bar{K}$\\
PSRO$_{\alpha\text{-Rank}}$ & $\alpha$-Rank & $\beta \in [0, 1]$ \& $K=\bar{K}$\\
PSRO$_{\text{CCE}}$ & Coarse Correlated & $\beta \in [0, 1]$ \& $K=\bar{K}$\\
\bottomrule
\end{tabular}
\caption{Specifications of PSRO variants. $\bar{K}$ is the number of updates needed to obtain a converged BR policy. The references for these methods: GDA~\protect\cite{fiez2021local}, InRL~\protect\cite{matignon2012independent}, PSRO$_{\text{P}}$~\protect\cite{ho1998iterated}, PSRO$_{\text{U}}$~\protect\cite{heinrich2016deep}, PSRO$_{\text{N}}$~\protect\cite{lanctot2017unified}, PSRO$_{\text{rN}}$~\protect\cite{balduzzi2019open}, PSRO$_{\alpha\text{-Rank}}$~\protect\cite{muller2020generalized}, PSRO$_{\text{CCE}}$~\protect\cite{marris2021multi}.}
\label{tab:gda_psro}
\end{table} 

\subsection{HPO Policy Optimization\label{sec:leader-problem}}

\begin{figure}[ht]
\centering
\includegraphics[width=0.75\columnwidth]{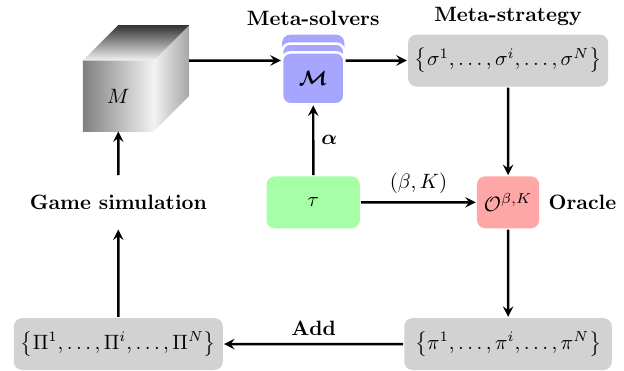}
\caption{Illustration of Self-adaptive Policy-Space Response Oracles (SPSRO). PSRO and PPSRO are two special cases of SPSRO. Illustration inspired by~\protect\cite{muller2020generalized}.}
\label{fig:psro}
\end{figure}

With the PPSRO introduced in the previous section, a natural problem is how to determine the values of $\bm{\alpha}$, $\beta$, and $K$ in PPSRO. To address this problem, we propose a novel algorithmic framework, Self-adaptive PSRO (SPSRO), which is shown in Figure~\ref{fig:psro} and Algorithm~\ref{alg:s-psro}. In the following, we first present the overall procedure of SPSRO, then define the performance
metric of a given selection of hyperparameter values, and finally formalize the hyperparameter optimization (HPO) problem where our objective is to learn an HPO policy that will self-adaptively select the optimal hyperparameter values of PPSRO during game solving.

Let $\tau \in \Gamma$ denote an HPO policy where $\Gamma$ is the policy space. Let $u^e = (\bm{\alpha}^e, \beta^e, K^e) \in \mathcal{U}$ denote the hyperparameter values selected according to the HPO policy $\tau$ at each epoch $e \ge 1$ of SPSRO, where $\mathcal{U}$ is the admissible set of the hyperparameter values. That is, we have $u^e \sim \tau$. We run one epoch of SPSRO as follows. (1) Compute the payoff tensor $M$ through game simulation (Line 3). (2) Compute the final joint meta-strategy $\bm{\sigma}^e$ (Line 4). (3) Expand each player's policy space (Line 5). Specifically, for each player $i$, we initialize the BR policy as $\pi^{i, \text{BR}, e} = \beta^e \pi^{i, \text{BR}, e-1} + (1 - \beta^e) \pi^{i, \text{random}}$, and then train the BR policy $\pi^{i, \text{BR}, e}$ for $K^e$ updates and add it to player $i$'s policy space $\Pi^i = \Pi^i \cup \{\pi^{i, \text{BR}, e}\}$. (4) Compute the performance metric $y^e(u^e)$ of the current selection (Line 6). (5) Select new hyperparameter values $u^{e+1}$ according to $\tau$ (Line 7).

\begin{algorithm}[ht]
\caption{SPSRO}
\label{alg:s-psro}
\begin{algorithmic}[1]
\STATE Initialize $\Pi^i$ with random policies, $\forall i \in \mathcal{N}$, $e \gets 1$, select initial hyperparameter values $u^1=(\bm{\alpha}^1, \beta^1, K^1)$, $\tau\in\Gamma$
\FOR{epoch $e\in\{1, 2, 3, \dots\}$}
\STATE Update payoff tensor $M$ via game simulation
\STATE Compute $\bm{\sigma}^e$ using $\bm{\mathcal{M}}$ and $\bm{\alpha}^e$: $\bm{\sigma}^e = \sum\nolimits_{b=1}^m \alpha_b^e \bm{\sigma}_b^e$
\STATE Expand policy spaces $\mathcal{O}^i$: $\Pi^i \gets \Pi^i \cup \mathcal{O}^i(\bm{\sigma}^e; \beta^e, K^e)$
\STATE Compute the performance metric $y^e(u^e)$
\STATE Select new hyperparameter values $u^{e+1}\sim\tau$
\ENDFOR
\end{algorithmic}
\end{algorithm}

Given a selection of the hyperparameter values $u^e$, we define its performance metric as $y^e(u^e) = \frac{\mathcal{R}(\bm{\sigma}^e)}{\mathcal{R}(\bm{\sigma}^{1})} + \frac{h^e}{h^{1}}$. Roughly speaking, it consists of two parts: the NashConv of all players and the BR training effort, which implies that using larger $K^e$ could obtain lower NashConv $\mathcal{R}(\bm{\sigma}^e)$ but at the cost of longer BR training time $h^e$, and using smaller $K^e$ could shorten the BR training time while at the cost of higher NashConv $\mathcal{R}(\bm{\sigma}^e)$. 

Given the performance metric, our objective is to learn an HPO policy $\tau$ by solving the following HPO problem: $\forall e \ge 1$,

\begin{equation}
    \arg\min\nolimits_{\tau \in \Gamma} y^e(u^e\sim\tau).\label{eq:leader-obj}
\end{equation} 

\section{A Novel Offline HPO Algorithm\label{sec:universal-hpo}}

The most straightforward method to optimize the HPO policy is to employ the classic HPO methods such as Bayesian optimization~\cite{snoek2012practical}. However, most HPO methods typically predict hyperparameter values based on online generated data (history of past epochs in the context of our work), and consequently, the performance could be poor. To overcome the limitations of online HPO, we propose a novel offline HPO method to optimize the HPO policy, which possesses the potential to transfer to different games without fine-tuning. 

\subsection{HPO as Sequence Modeling}

\begin{figure}[ht]
\centering
\includegraphics[width=0.9\columnwidth]{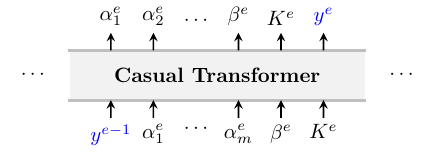}
\caption{HPO based on Transformer. At each epoch, the Transformer model predicts parameter values in an autoregressive manner using a causal self-attention mask, i.e., each predicted parameter value will be fed into the model to generate the next parameter value.}
\label{fig:hpo-transformer}
\end{figure}

As presented in Algorithm~\ref{alg:s-psro}, selecting the hyperparameter values in SPSRO can be naturally regarded as a sequence modeling problem where we model the probability of the next token $x^e$ conditioned on all prior tokens: $P_{\theta}(x^e \vert x^{<e})$, similar to the decoder-only sequence models~\cite{zhao2023survey,touvron2023llama}. Specifically, we consider the sequence of hyperparameter values up to $e$-th epoch:
\begin{align}
    \mathcal{H}^e &= (\cdots, \alpha_1^e, \cdots, \alpha_m^e, \beta^e, K^e, y^e). \label{eq:token-sequence-part}
\end{align}
Figure~\ref{fig:hpo-transformer} presents the overview of the architecture. 

\subsection{Tokenization} 
We convert each element in Eq.~(\ref{eq:token-sequence-part}) into a single token. The idea is to normalize and discretize each element such that it falls into one of $Q$ bins each with size 1. $Q$ is referred to as the quantization level. Specifically, we have:
\begin{equation}
   \bar{x}^e = \text{int}[x_{\text{norm}}^e \cdot Q],
\end{equation}
where $x_{\text{norm}}^e = (x^e - x_{\text{min}}) / (x_{\text{max}} - x_{\text{min}})$. $x_{\text{min}}$ and $x_{\text{max}}$ are determined by the space $\mathcal{U}$ and the range of $y^e$ is determined by observed values in the offline dataset (introduced in the next subsection) or the underlying games (e.g., normal-form and extensive-form). After tokenization, we have:
\begin{align}
    \bar{\mathcal{H}}^e &= (\cdots, \bar{\alpha}_1^e, \cdots, \bar{\alpha}_m^e, \bar{\beta}^e, \bar{K}^e, \bar{y}^e).\label{eq:converted-token-sequence-part}
\end{align}

\subsection{Training Dataset\label{sec:train_dataset}} 
To train the Transformer model $\theta$, one of the critical steps is to generate the offline dataset, which is non-trivial due to the computational complexity of running SPSRO. Specifically, to generate a dataset $\mathcal{D}$ consisting of $\vert \mathcal{D} \vert$ sequences of Eq.~(\ref{eq:token-sequence-part}) or Eq.~(\ref{eq:converted-token-sequence-part}), we need to run the SPSRO for $\vert \mathcal{D} \vert$ times. However, it is well-known that running PSRO can be computationally difficult in complex games. One of the main difficulties is that solving the meta-game using $\alpha$-Rank could be NP-hard~\cite{yang2020alphaalpha} and as it requires enumerating all the joint strategies to construct the response graph~\cite{omidshafiei2019alpha}, it is time-consuming as the progress of the PSRO procedure. To more efficiently generate the training dataset, we use a simple pruning technique to constrain the size of the support set of the meta-strategy when using $\alpha$-Rank to solve the meta-game in each epoch of the PSRO algorithm.

Let $C$ denote the maximum size of the support set of the meta-strategy. At the first $C$ epochs, we follow the standard PSRO$_{\alpha\text{-Rank}}$. After that, at each epoch $e > C$, we construct the meta-game of size $(C+1) \times (C+1)$ where a new BR policy is obtained by deep RL algorithms such as DQN~\cite{mnih2015human}. Next, we compute the meta-distribution by solving the meta-game through $\alpha$-Rank. Let $\pi^i_{\text{min}}$ denote the policy with the minimum probability in the meta-distribution. Then, we get the final meta-strategy by setting $\sigma^{i, e}_{\alpha\text{-Rank}}(\pi^i_{\text{min}}) = 0$ and normalizing the resulting distribution. 

During the dataset generation, we employ the widely used tool, Optuna~\cite{akiba2019optuna}, to determine the value of $u^e$ at each epoch $e$. Using the terminology of offline RL~\cite{levine2020offline,chen2021decision}, Optuna is a \textit{behavior policy} to generate the offline training dataset (see Appendix B for an example code of Optuna showing how to generate the dataset). Furthermore, we distinguish between normal-form games (NFGs) and extensive-form games (EFGs) when generating the dataset. The primary reason is that the parameters of interest are different. In EFGs, in addition to the weights of different meta-solvers $\bm{\alpha}$, Eq. (\ref{eq:token-sequence-part}) also consists of the parameters related to the BR oracle, $\beta$ and $K$. Therefore, the transformer model trained on the NFG dataset cannot be directly applied to EFGs. In addition, generating the NFG dataset is relatively easier as the computational difficulty mainly resulted from the meta-game solving using $\alpha$-Rank, which can be effectively addressed by the previously proposed pruning technique. For the EFG dataset, obtaining the BR policies requires extra computational overhead as the BR policies are typically approximated via deep RL algorithms such as DQN~\cite{mnih2015human}.

\begin{figure*}[ht]
\centering
\includegraphics[width=\textwidth]{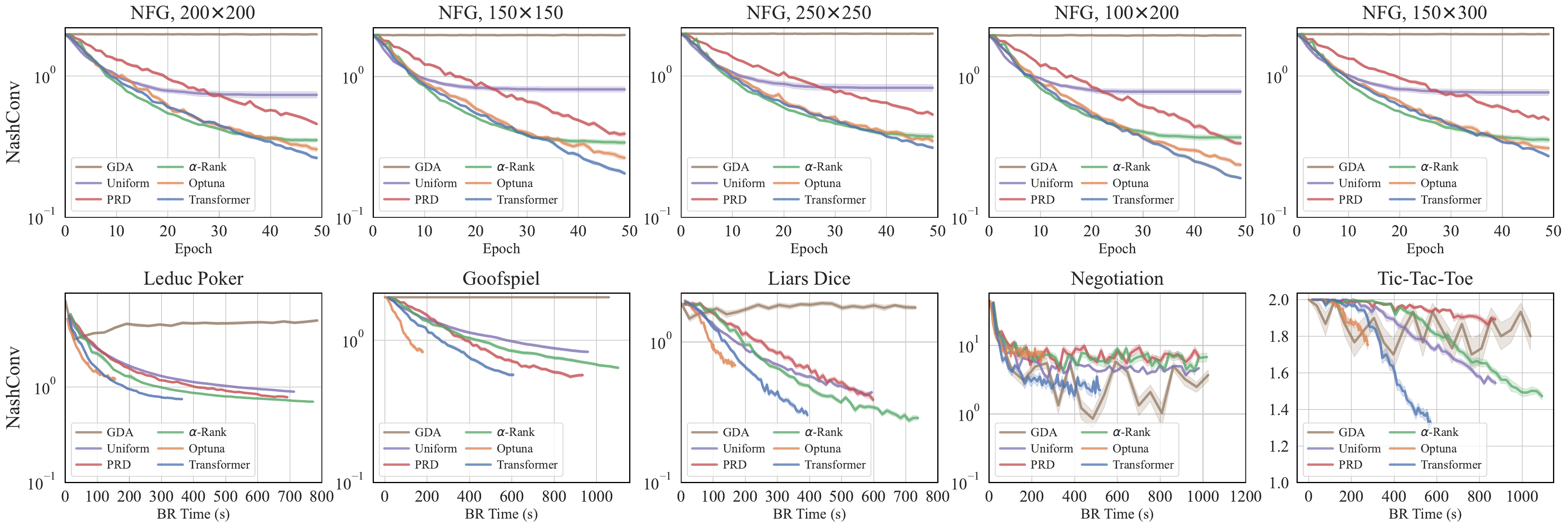}
\caption{Evaluation performance. The top and bottom rows correspond to NFGs and EFGs, respectively.}
\label{fig:results-summary}
\end{figure*}

\subsection{Loss Function and Inference} 

Given the dataset $\mathcal{D}$, we train the Transformer model $\theta$ by maximizing the log-likelihood for each sequence $\mathcal{H}^{\bar{e}} \sim \mathcal{D}$:
\begin{equation}
    \mathcal{L}(\theta; \mathcal{H}^{\bar{e}}) = \sum\nolimits_{n=1}^{\bar{e} (m + 3)} \log P_{\theta}(\bar{\mathcal{H}}^n \vert \bar{\mathcal{H}}^{1:n-1}),
\end{equation}
where $\bar{e}$ is the maximum number of epochs, $\bar{\mathcal{H}}^n$ is the $n$-th token in Eq.~(\ref{eq:converted-token-sequence-part}), and $\bar{\mathcal{H}}^{1:n-1}$ is all tokens up to the $(n-1)$-th token in Eq.~(\ref{eq:converted-token-sequence-part}). After training, we can apply the Transformer $\theta$ to a given game to predict the value of $u^e$. Specifically, we reverse the tokenization to obtain the token distribution:
\begin{equation}
    p_{\theta}(x \vert \cdot) = \frac{Q \cdot P_{\theta}(\bar{x} \vert \cdot)}{(x_{\text{max}} - x_{\text{min}})}.
\end{equation}
Then, we can sample $u^e$ from the model’s prior distribution and thus, define the HPO policy as follows:
\begin{equation}
\begin{aligned}
    \tau(u^e \vert \mathcal{H}^{e-1}) = & \prod\nolimits_{b=1}^{m} p_{\theta}(\alpha_b^e \vert \mathcal{H}^{e-1}, \alpha_1^e, \cdots, \alpha_{b-1}^e) \\
    & \times p_{\theta}(\beta^e \vert \mathcal{H}^{e-1}, \{\alpha_b^e\}_{1 \leq b \leq m}) \\
    & \times p_{\theta}(K^e \vert \mathcal{H}^{e-1}, \{\alpha_b^e\}_{1 \leq b \leq m}, \beta^e).
\end{aligned}
\end{equation}
That is, at epoch $e$, $\tau$ predicts each parameter value in $u^e$ conditioned on: i) the sequence of past epochs $\mathcal{H}^{e-1}$, and ii) the values of the preceding predicted parameters.

\section{Experiments\label{sec:experiments}}

In this section, we evaluate the effectiveness of SPSRO.

\subsection{Experimental Setup}

\begin{figure*}[ht]
\centering
\includegraphics[width=\textwidth]{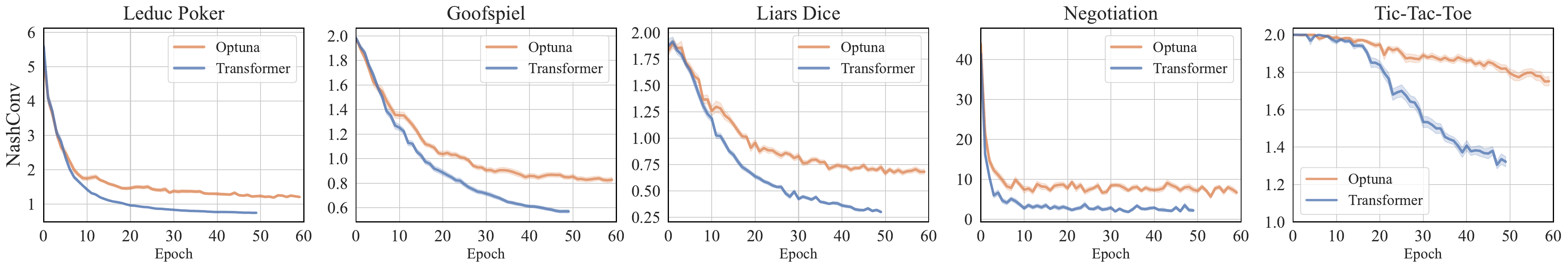}
\caption{NashConvs of Optuna and Transformer in different extensive-form games.}
\label{fig:nashconv-optuna-transformer}
\end{figure*}

All experiments are performed on a machine with a 24-core 3.2GHz Intel i9-12900K CPU and an NVIDIA RTX 3060 GPU. All the results are averaged over 30 independent runs. More experimental details can be found in Appendix C.

\noindent\textbf{Games}. We consider the following games. (1) Normal-form games (NFGs) of size $\vert \mathcal{A}_1 \vert \times \vert \mathcal{A}_2 \vert$. The payoff matrices are randomly sampled from the range $[-1, 1]$. The set of size is $\{150\times150, 200\times200, 250\times250, 100\times200, 150\times300\}$. (2) Extensive-form games (EFGs): Leduc, Goofspiel, Liar's Dice, Negotiation, and Tic-Tac-Toe, which are implemented in OpenSpiel~\cite{lanctot2019openspiel}. Please refer to Section~\ref{sec:train_dataset} for the reasons to distinguish between NFGs and EFGs. 

\noindent\textbf{Methods}. (1) GDA~\cite{fiez2021local}. At each epoch, a player only best responds to the opponent's newest BR action (NFGs) or policy (EFGs). (2) Uniform~\cite{heinrich2016deep}. The meta-distribution is the uniform distribution. (3) PRD~\cite{lanctot2017unified,muller2020generalized}, an approximation of Nash equilibrium. We choose PRD instead of an exact Nash solver as it has been widely adopted in PSRO-related research. (4) $\alpha$-Rank~\cite{muller2020generalized}. (5) Optuna~\cite{akiba2019optuna}. The parameter values are determined by Optuna. (6) Transformer. The parameter values are determined by the Transformer model. Among these methods, (2) to (4) are classic PSRO methods that only involve a single meta-solver, while (5) and (6) are SPSRO methods that involve multiple meta-solvers and use different HPO approaches.

\noindent\textbf{Training and Testing.} We generate the training datasets for NFGs and EFGs separately. For NFGs, we generate the dataset on the game of size $\vert \mathcal{A}_1 \vert \times \vert \mathcal{A}_2 \vert = 200 \times 200$. For EFGs, we generate the dataset on the Leduc Poker. During testing, in addition to the games used to generate the dataset, we directly apply the trained Transformer model to the other games to verify the zero-shot generalization ability of the model.

\subsection{Results}

The results are summarized in Figure~\ref{fig:results-summary}. From the results, we can draw several conclusions as follows.

\textit{By combining multiple meta-solvers, we could obtain better performance than using a single meta-solver.} In all the NFGs, the final NashConvs of Optuna and Transformer are lower than that of the classic PSRO baselines considering a single meta-solver (Uniform, PRD, or $\alpha$-Rank). For EFGs, in Negotiation and Tic-Tac-Toe, the final NashConvs of Transformer are lower than the classic PSRO baselines. The results clearly verify the necessity of synergistically integrating multiple meta-solvers during game solving.

\textit{Transformer-based HPO could achieve better performance.} Transformer-based HPO can learn a better prior distribution of hyperparameter values from offline data, providing a better scheme for weighting multiple meta-solvers, and therefore, achieving better performance than Optuna which is an online method and only relies on past epochs to obtain the prior distribution of hyperparameter values. In addition, in EFGs, we found that the NashConv of Optuna decreases quickly, but converges to a high value (also shown in Figure~\ref{fig:nashconv-optuna-transformer}). In contrast, the Transformer can converge to a lower NashConv, though it needs a longer time in terms of BR training. 

\textit{Transformer has the potential to provide a universal and plug-and-play hyperparameter value selector.} As shown in Figure~\ref{fig:results-summary}, the trained Transformer model can be applied to the games that are different from the training dataset: for NFGs, it can be applied to games with different sizes of action (strategy) spaces, and for EFGs, it can be applied to different games even with different reward scale (e.g., the maximum reward in Goofspiel is 1 while in Negotiation it is 10). This corresponds to the desiderata of a universal and plug-and-play hyperparameter value selector as mentioned in Section~\ref{sec:universal-hpo}.

\textit{Given a set of meta-solvers, none of them can consistently beat (dominate) all the others during game solving (the observation in Section~\ref{sec:motivating-example}).} This can be derived by comparing the performance of the three single-solver-based PSRO baselines: Uniform, PRD, and $\alpha$-Rank. For example, consider the NFG of size $200 \times 200$. In the early stage of PSRO, the NashConv of Uniform is lower than PRD and $\alpha$-Rank. In the middle stage, $\alpha$-Rank quickly surpasses Uniform and PRD, but Uniform still performs better than PRD. However, in the final stage, Uniform is beaten by PRD and $\alpha$-Rank. Moreover, we note that in the final stage, PRD could also perform better than $\alpha$-Rank, as shown in the NFG of size $100 \times 200$. Similar results are observed in EFGs, further verifying the conclusion.

For EFGs, in Figure~\ref{fig:results-summary}, at first glance, one may come to the conclusion that Optuna is a better option than Transformer. However, we note that the $x$-axis in Figure~\ref{fig:results-summary} is the BR running time (which is appropriate as in our experiments only one BR policy is added to each player's policy space). To avoid this misleading conclusion, we plot the NashConv versus epoch in Figure~\ref{fig:nashconv-optuna-transformer}. The results clearly show that instead of terminating too early, Optuna cannot further decrease the NashConv even if it is given more epochs (10 epochs more than Transformer). The primary reason we hypothesize is that, as the SPSRO progresses, it is a struggle for Optuna to balance the two parts (NashConv and BR training time) in the performance metric $y^e$. Specifically, at the latter stage of SPSRO, the second term in $y^e$ would dominate the first term, even though they have been normalized by using the values obtained at the first epoch, making Optuna suggest a smaller number of updates for the BR policies (see Figure~\ref{fig:br-time} in Section~\ref{sec:more_discussion}), which on the contrary cannot further decrease the NashConv because the quality of the BR policies would be low without providing enough training amount. The results demonstrate that Transformer is more effective in handling such a dilemma.

\subsection{More Discussion\label{sec:more_discussion}}

In this section and Appendix C, we provide more discussion to further deepen our understanding of our approach.

\begin{figure}[ht]
\centering
\includegraphics[width=0.49\columnwidth]{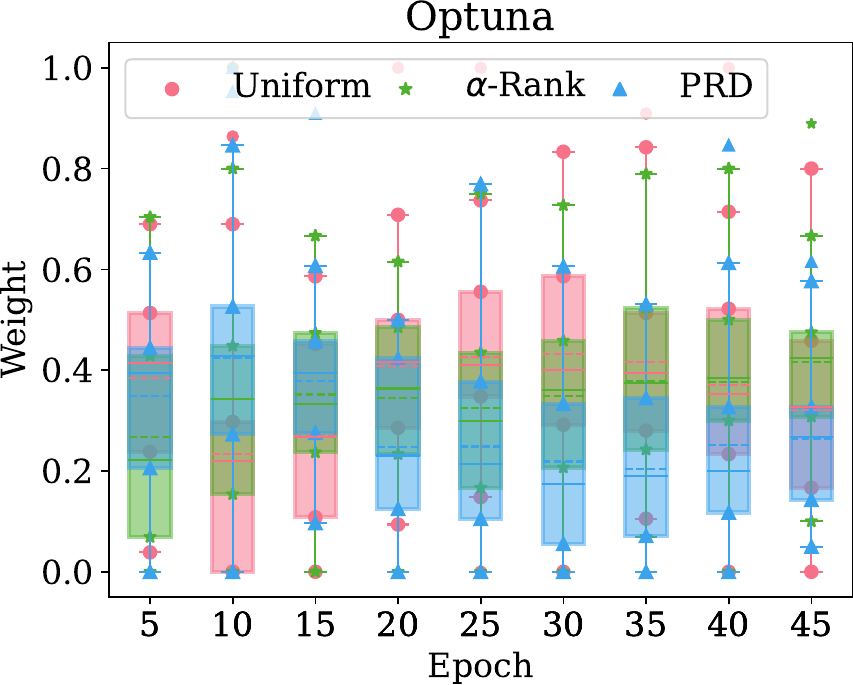}
\includegraphics[width=0.49\columnwidth]{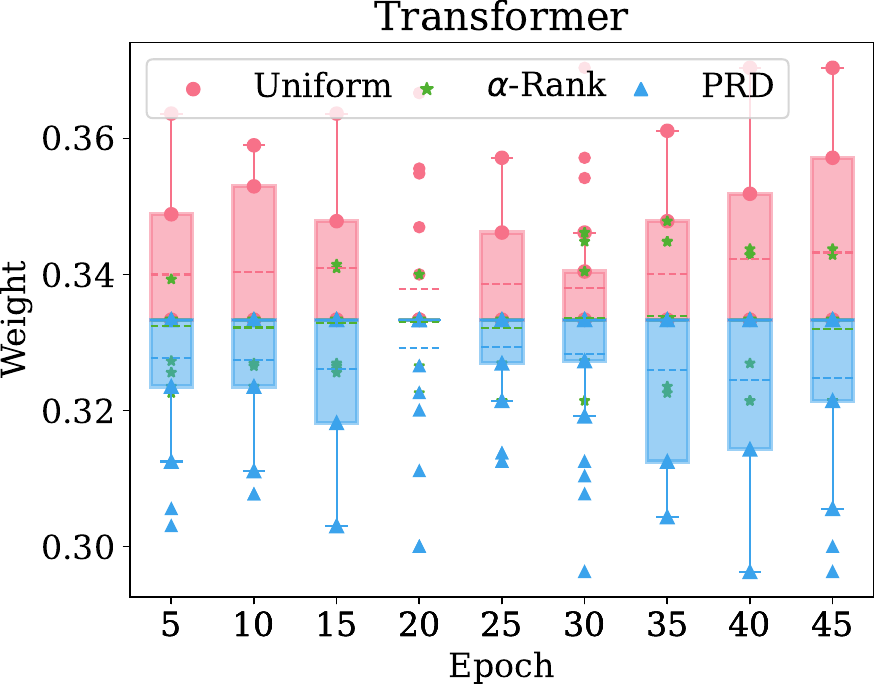}
\caption{Weights of three meta-solvers determined by Optuna and Transformer in the NFG of size 200$\times$200.}
\label{fig:weight-vectors}
\end{figure}

Figure~\ref{fig:weight-vectors} shows the weights of different meta-solvers (Uniform, PRD, and $\alpha$-Rank) determined by Optuna and Transformer during SPSRO running. We can see that the weights determined by Optuna vary dramatically throughout SPSRO running, while Transformer's predictions are more stable (around $1/3$ for each solver). We hypothesize that such a relatively stable weighting scheme for multiple meta-solvers is necessary to obtain better performance. In addition, an interesting observation is that the weights of Uniform and PRD change almost in a mirror form. More results can be found in Appendix C.

\begin{figure}[ht]
\centering
\includegraphics[width=0.49\columnwidth]{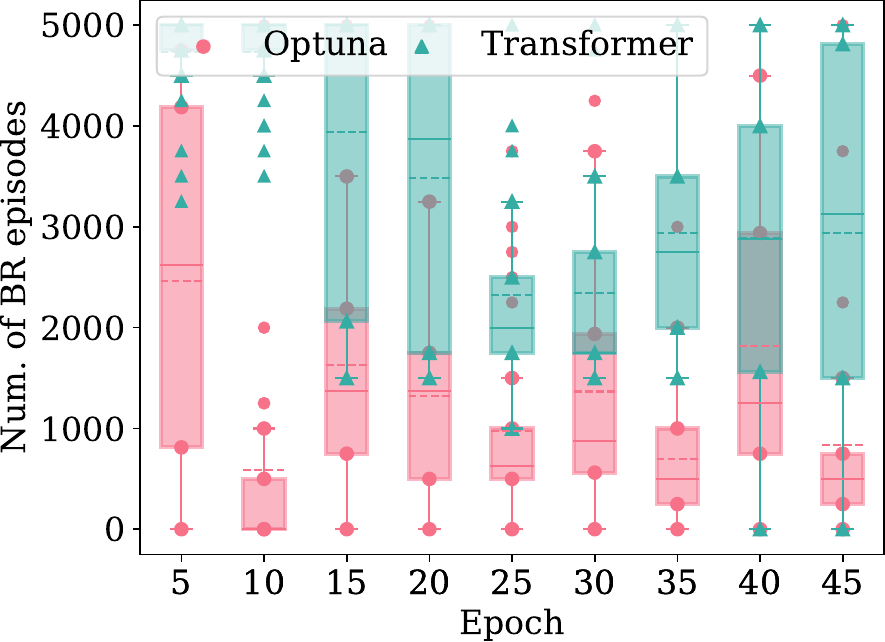}
\includegraphics[width=0.49\columnwidth]{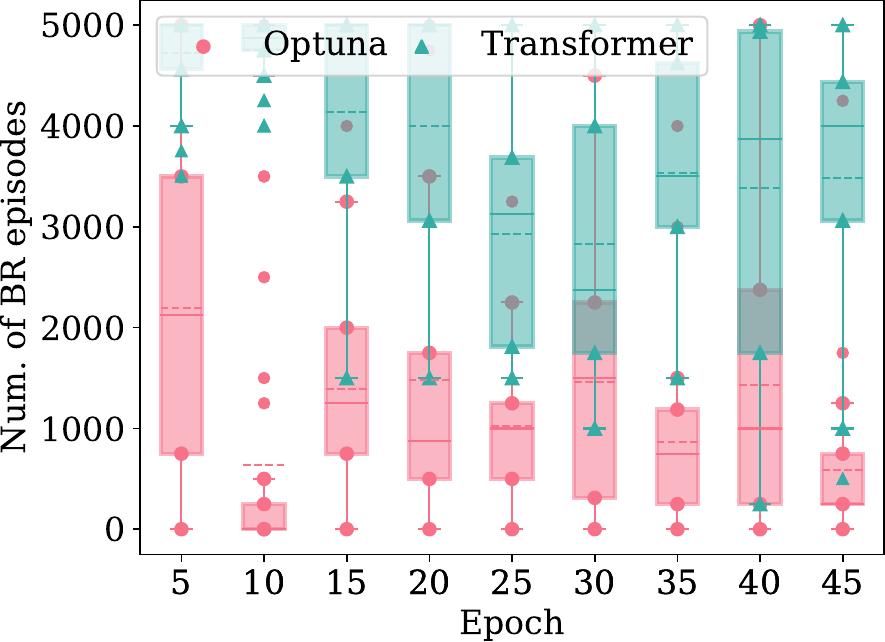}
\caption{Numbers of BR training episodes determined by Optuna and Transformer in EFGs. (Left) Leduc. (Right) Goofspiel.}
\label{fig:br-time}
\end{figure}

In Figure~\ref{fig:br-time}, we found that Optuna and Transformer follow very different patterns to select the computing amount used for training the BR policy at each epoch. For Optuna, the number of episodes suddenly decreases to a very low value after about 10 epochs. Intuitively, when $K$ is much smaller than $\bar{K}$ (the maximum number of episodes to obtain a converged BR policy), the policy obtained through the BR oracle may be far away from the true BR policy, resulting in poor performance. This is also reflected in Figure~\ref{fig:nashconv-optuna-transformer} where Optuna cannot obtain a lower NashConv even if it is given more epochs. In contrast, by offline learning, Transformer could better trade-off between the NashConv and BR training time and hence, performs better than Optuna. More results can be found in Appendix C.

\section{Conclusions\label{sec:conclusion}}

In this work, we first attempt to explore the possibility of self-adaptively determining the optimal hyperparameter values in the PSRO framework and provide three contributions: (1) the parametric PSRO (PPSRO) which unifies GDA and various PSRO variants; (2) the self-adaptive PSRO (SPSRO) where we aim to learn a self-adaptive HPO policy; (3) a novel offline HPO approach to optimize the HPO policy based on the Transformer architecture. The well-trained Transformer-based HPO policy has the potential of transferring to different games without fine-tuning. Experiments on different games demonstrate the superiority of our approach over different baselines.

\bibliographystyle{named}
\bibliography{ijcai24}

\clearpage
\appendix

\section{Frequently Asked Questions\label{app:fqa}}

\textbf{Q1. Why self-adaptive HPO policy is necessary in PSRO?} 

PSRO~\cite{lanctot2017unified} since pioneered has been the main algorithm for solving various games. However, determining the hyperparameter values in PSRO could be difficult in complex games. On one hand, at each epoch, it needs to solve the meta-game to obtain the meta-strategies of all players, which could be computationally difficult, e.g., solving $\alpha$-Rank may be NP-hard~\cite{yang2020alphaalpha}. On the other hand, at each epoch, it needs to compute the BRs of all players, which could be time-consuming because, in complex games such as poker and soccer, the BRs are often approximated by deep RL algorithms such as DQN~\cite{mnih2015human} and obtaining good approximated BRs will require a large number of updates. Moreover, existing works have suggested various meta-solvers such as Uniform~\cite{heinrich2016deep}, Nash~\cite{lanctot2017unified}, $\alpha$-Rank~\cite{muller2020generalized}, correlated equilibrium~\cite{marris2021multi}, and neural meta-solver~\cite{feng2021neural}. However, there is no explicit rule of which one is the best in terms of learning performance.

Therefore, there is a challenging dilemma: (1) we must determine these hyperparameter values before running PSRO, and (2) evaluating a particular setting is hard because the performance is unknown until the end of the PSRO running due to its nature of iterative policy-space expansion. Thus, instantiating and running PSRO typically involve extensive domain knowledge, rendering it unfriendly to researchers, especially those who are not familiar with game theory. 

To further boost PSRO-based research and make it more applicable to real-world problems, a natural idea is to establish a hyperparameter value selector that can be employed to self-adaptively determine the optimal hyperparameter values for different games (even if the games are very different from each other such as normal-form games versus extensive-form games) with or without fine-tuning,  which is exactly the motivation of this work. Ideally, such a self-adaptive hyperparameter value selector can help reduce the effort required for researchers to conduct the costly hyperparameter tuning when applying PSRO to different games, and hence, we believe it will benefit the game community.

\noindent\textbf{Q2. More explanations on mixing multiple meta-solvers.}

As PSRO is a general algorithmic framework, any meta-solver can be used to compute the meta-distribution, though different meta-solvers are designed to optimize toward different solution concepts. However, it is worth noting that the objective that PSRO achieves is primarily determined by the evaluation metric. For example, in~\cite{muller2020generalized}, even if $\alpha$-Rank is chosen as the meta-solver, one can still evaluate the convergence of PSRO to Nash equilibrium by computing the NashConv, the metric to measure the distance between the current policy to Nash equilibrium. In this sense, we note that the observations in Section 4 hold in terms of NashConv, which supports our motivation of combining multiple meta-solvers during game solving using PSRO. Moreover, as mentioned in Section 4, our observations do not cause inconsistency with~\cite{wang2022evaluating} as we just evaluate PSRO as an online algorithm. Finally, we note that combining multiple meta-solvers or changing from one meta-solver to another has been explored and demonstrated better performance in PSRO~\cite{feng2021neural} and double oracle (DO)~\cite{wang2019deep}, which provides support to our motivation.

\noindent\textbf{Q3. Discussion on the hyperparameters for the BR oracle.}

In this work, we consider two hyperparameters in the BR oracle: the initialization parameter $\beta$ and the number of BR policy updates $K$. Here, we provide more explanations on the parameter $K$. (1) PSRO is a deep learning extension of double oracle (DO), where RL is adopted to compute the best response. However, RL is not guaranteed to compute the exact best response~\cite{lanctot2017unified}, a more suitable notion is \enquote{better response}, i.e., a new response better than the existing responses, which is also considered in DO~\cite{jain2013security}. In this sense, $K$ is a hyperparameter determining the amount of computing used to learn the better response. (2) Although different methods can be adopted to learn the BR policy, e.g., DQN and PPO, it is essential to use the same method (DQN in this work) to ensure a fair comparison when comparing different PSRO methods. In other words, the comparison between PSRO$_{\text{N}}$-DQN and PSRO$_{\text{N}}$-PPO would not induce any valid argument \textit{on the parameter} $K$. Furthermore, even for the same BR learning method, keeping the same configuration for all the other hyperparameters (e.g., batch size) is also indispensable. (3) As this is the first attempt to explore the possibility of self-adaptively determining the optimal hyperparameter values in the PSRO framework, we only focus on some of the hyperparameters. As for self-adaptively determining more hyperparameters such as learning rate and batch size, more techniques may be required~\cite{cheny2022towards,zahavy2020self} and we leave for future works.

\noindent\textbf{Q4. More explanations on the NashConv values.}

In some EFGs, the NashConv values seem relatively high, e.g., in Tic-Tac-Toe. Nevertheless, we note that this does not violate the goal achieved by our framework: achieve a better balance between minimizing the NashConv of all players and the BR training effort. In this sense, Transformer performs better than Optuna (as well as all other baselines). Moreover, though not fully converged, from Figure 5 and Figure 6 in the main text we can see that, as the learning process progresses, the NashConv of Transformer shows a continuously and relatively quickly decreasing trend, while the NashConv of Optuna decreases much more slowly.

\section{More Related Works\label{app:more-related-works}}
 
Transformers~\cite{vaswani2017attention} have demonstrated outstanding performance in various real-world tasks including natural language processing (NLP)~\cite{vaswani2017attention,beltagy2020longformer}, computer vision (CV)~\cite{liu2021swin,yuan2021tokens}, and 3D point cloud~\cite{zhao2021point}. Transformers have also shown the ability of symbolic manipulation~\cite{lample2019deep,chenmark2021evaluating,li2022competition}, numerical manipulation~\cite{charton2021linear,muller2022transformers}, or both~\cite{cheny2022towards}. Recently, Transformers have also been applied to offline RL and achieved state-of-the-art performance in various RL benchmarks~\cite{chen2021decision,janner2021offline,zheng2022online,xu2022prompting,lee2022multigame,wen2022multi}. In this work, it is natural to model the HPO policy learning as a sequence modeling problem and thus, Transformers can be used to predict hyperparameter values. Intuitively, we aim to learn an HPO policy that self-adaptively determines \textit{a sequence of choices of the hyperparameter values} to optimize the objective, which is similar to the RL problem where an agent tries to maximize his return through \textit{a sequence of actions}. Therefore, we use the Decision Transformer~\cite{chen2021decision} to predict the hyperparameter values in SPSRO.

Our work is also related to empirical game theoretic analysis (EGTA), an empirical methodology that bridges the gap between game theory and simulation for practical strategic reasoning~\cite{wellman2006methods}. In EGTA, an empirical game is estimated via simulation over the combinations of a set of strategies and then can be analyzed with standard methods. In iterative approaches to EGTA, the game model is extended by iteratively generating and adding new strategies to the strategy spaces of players. One of the most challenging problems in EGTA is the strategy exploration problem (how to direct the iterative strategy generation process to construct effective game models with minimal iteration) and a set of works have been done toward investigating this problem~\cite{schvartzman2009stronger,schvartzman2009exploring,jordan2010strategy,wang2022evaluating}. In~\cite{schvartzman2009stronger}, tabular RL has been adopted to serve as the BR oracle to generate new strategies. Investigation of strategy exploration was advanced significantly by the introduction of PSRO~\cite{lanctot2017unified} which is a flexible framework for iterative EGTA, where at each epoch, new strategies are generated through deep reinforcement learning (DRL) algorithm such as DQN~\cite{mnih2015human}.

\section{More Experimental Details\label{app:exp-details}}

In this section, we provide more experimental details.

\noindent\textbf{Dataset.} We separately generate the datasets for normal-form games and extensive-form games. For the normal-form game, we generate the dataset on the game of size $\vert \mathcal{A}_1 \vert \times \vert \mathcal{A}_2 \vert = 200 \times 200$, and the dataset contains $1000$ sequences with the form of Eq. (4) with $\bar{e} = 50$. For the extensive-form game, we generate the dataset on the Leduc Poker, and the dataset consists of a total of $1359$ sequences with the form of Eq. (4) with $\bar{e} = 50$. During dataset generation, we use Optuna~\cite{akiba2019optuna} as the \textit{behavior policy}. Using the terminology of offline RL~\cite{levine2020offline,chen2021decision,lee2022multigame}, each sequence is a \textit{trajectory} recording the process of SPSRO under the behavior policy. To more clearly show how to generate the dataset, we present an example code in Figure~\ref{fig:optuna_flow}. At each epoch $e$, Optuna first suggests new values for the hyperparameters (Line 6--10), including the weights of multiple meta-solvers and the hyperparameters related to the BR oracle, then runs one epoch of PSRO and computes the metric $y^e$ (Line 13--15). After $\bar{e}$ epochs, Optuna returns the sequence of Eq. (4) (Line 17--20). This process will be repeated for $\vert \mathcal{D} \vert$ times to generate the training dataset.

\setcounter{figure}{8}
\begin{figure}[ht]
\centering
\includegraphics[width=\columnwidth]{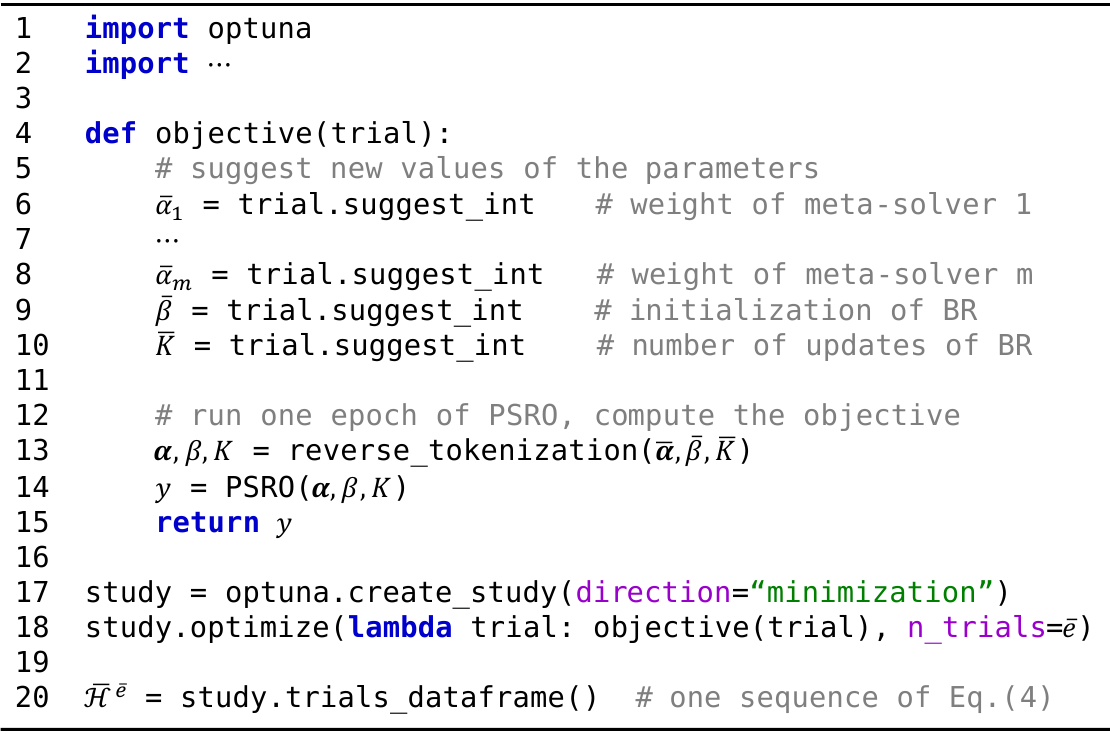}
\caption{An example code for generating one sequence of Eq. (4).}
\label{fig:optuna_flow}
\end{figure}

\noindent\textbf{Model Architecture.} We utilize Decision Transformer~\cite{chen2021decision} as our underlying model architecture. Specifically, 4 linear layers are used to project the weights of the three meta-solvers (Uniform, PRD, and $\alpha$-Rank) $\bm{\alpha}^e$, the initialization parameter value $\beta^e$, the number of BR episodes $K^e$, and the objective $y^e$ to the token embeddings which will be fed into a Transformer block followed by layer normalization~\cite{ba2016layer}. The outputs are then mapped by a linear layer to the logits which will be used to compute the loss during training or infer the next token during inference. Moreover, an embedding for each epoch is learned and added to each token embedding. More details can be found in~\cite{chen2021decision} and the corresponding codebase\footnote{https://github.com/kzl/decision-transformer}.

\noindent\textbf{Hyperparameters.} The hyperparameters are provided in Table~\ref{tab:hyperparameters} where those of the Decision Transformer are similar to~\cite{chen2021decision}. We give some remarks on some of the hyperparameters as follows. (1) The \textit{context length} used in the Transformer context corresponds exactly to the \textit{maximum number of epochs} of SPSRO. (2) The \textit{return-to-go} is simply set to $0$ and the operation of decreasing return-to-go with reward is removed. In other words, the objective value at each intermediate epoch is only used to serve as a signal to guide the selection of new hyperparameter values for the next epoch. (3) In this work, we consider three meta-solvers: Uniform, PRD, and $\alpha$-Rank ($m=3$), the most commonly used ones in most of the existing works. As for considering more meta-solvers such as correlated equilibrium and neural meta-solver, we leave it for future works.

\begin{table}[ht]
\setlength\tabcolsep{5pt}
\begin{tabular}{l|l}
\toprule
\textbf{Hyperparameter} & \textbf{Value} \\
\midrule
Number of Transformer blocks & $2$ \\
Number of attention heads & $4$ \\
Embedding dimension & $128$ \\
Batch size & $64$ \\
Context length (i.e., $\bar{e}$) & $50$ \\
Return-to-go & $0$ \\
Nonlinearity & GeLU \\
Transformer training epochs & $50$ \\
Dropout & $0.1$ \\
Learning rate & $3 * 10^{-4}$ \\
Adam betas & $(0.9, 0.95)$ \\
Grad norm clip & $1.0$ \\
Weight decay & $0.1$ \\
Warmup tokens & $5000$ \\
Final tokens & \makecell[l]{NFG: $2 * 50 * 1000 * 4$\\EFG: $2 * 50 * 1359 * 6$}\\
\midrule
Quantization level $Q$ & 20 \\
Number of meta-solvers $m$ & 3 \\
Size of the support set C & 10 \\
Max number of BR episodes $\bar{K}$ & 5000 \\
\bottomrule
\end{tabular}
\caption{Hyperparameters.}
\label{tab:hyperparameters}
\end{table}

\begin{figure}[hbt!]
    \centering
    \includegraphics[width=0.49\columnwidth]{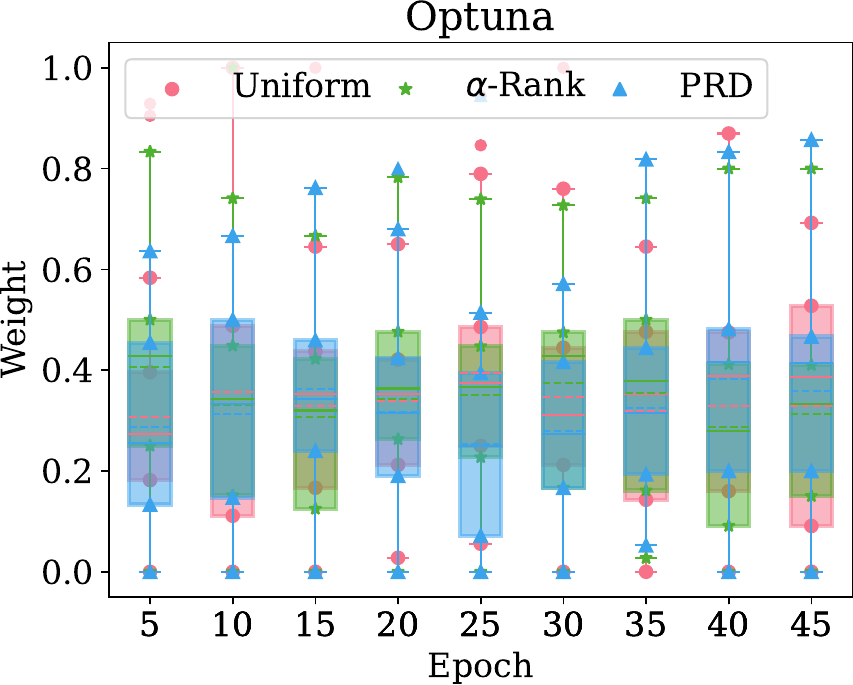}
    \includegraphics[width=0.49\columnwidth]{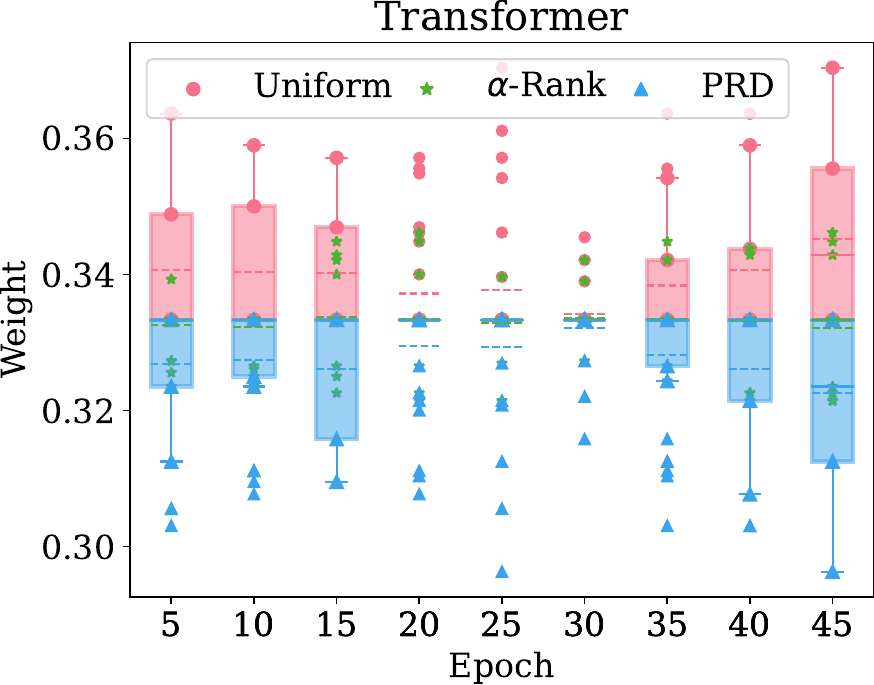}
    \caption{Weights of meta-solvers in \textbf{NFG 150$\times$150}.}
    \label{fig:nfg_150_150}
\end{figure}

\begin{figure}[hbt!]
    \centering
    \includegraphics[width=0.49\columnwidth]{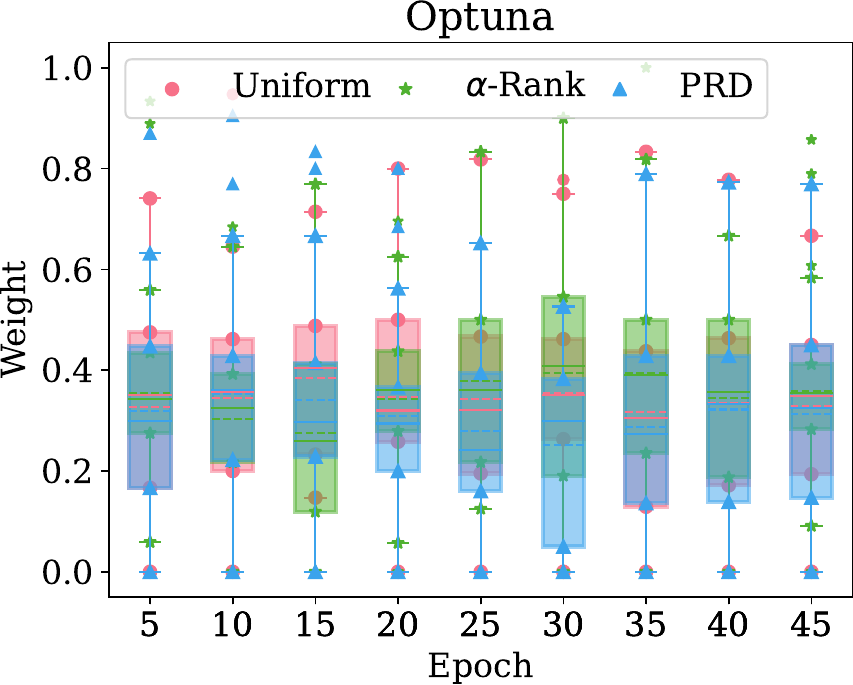}
    \includegraphics[width=0.49\columnwidth]{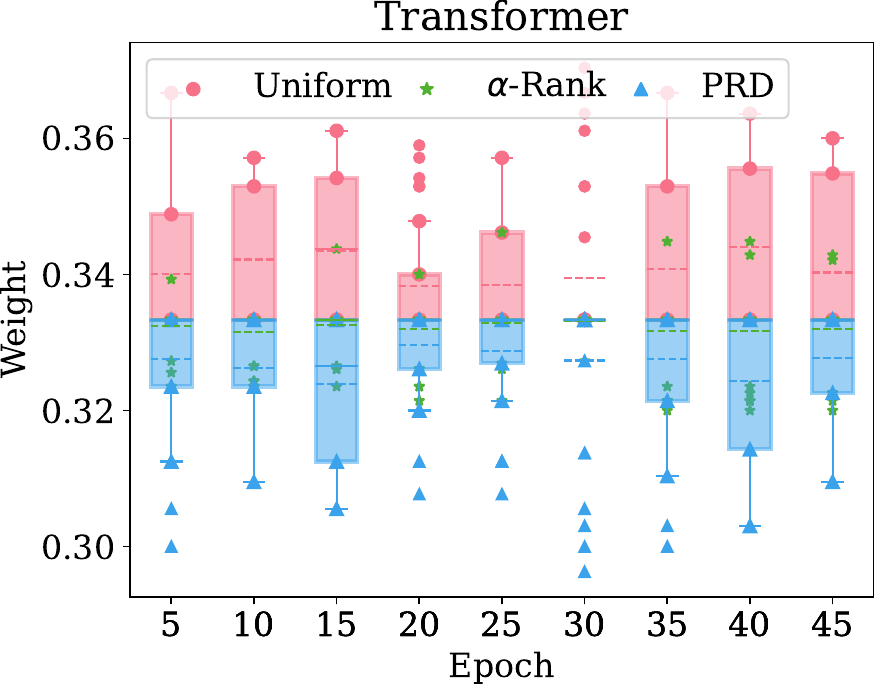}
    \caption{Weights of meta-solvers in \textbf{NFG 250$\times$250}.}
    \label{fig:nfg_250_250}
\end{figure}

\begin{figure}[hbt!]
    \centering
    \includegraphics[width=0.49\columnwidth]{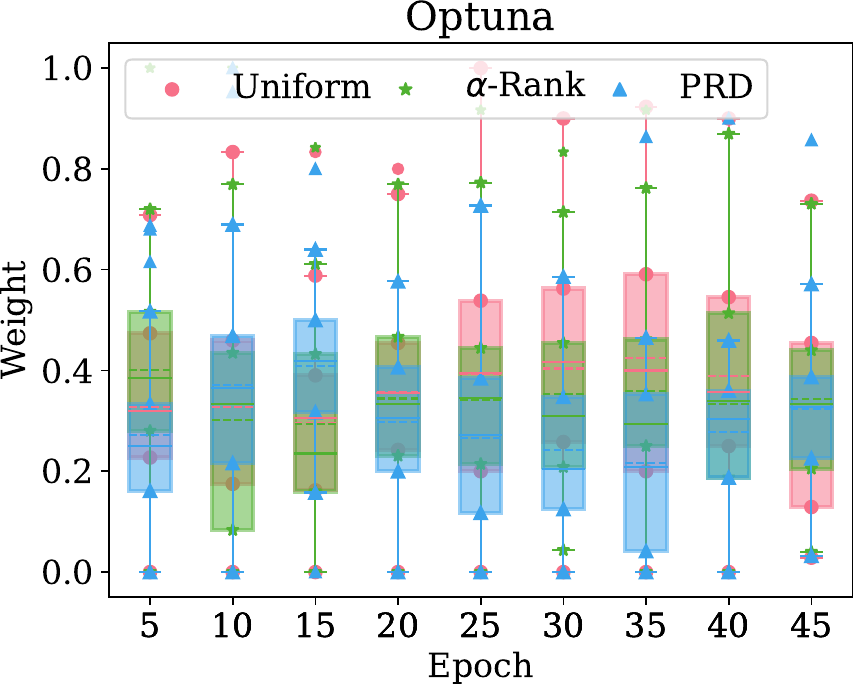}
    \includegraphics[width=0.49\columnwidth]{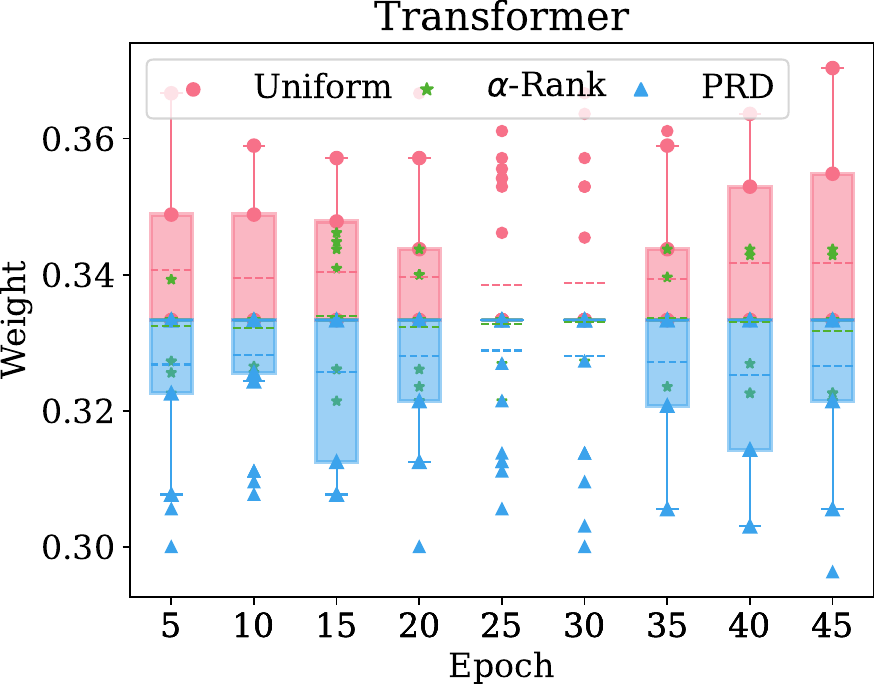}
    \caption{Weights of meta-solvers in \textbf{NFG 100$\times$200}.}
    \label{fig:nfg_100_200}
\end{figure}

\begin{figure}[hbt!]
    \centering
    \includegraphics[width=0.49\columnwidth]{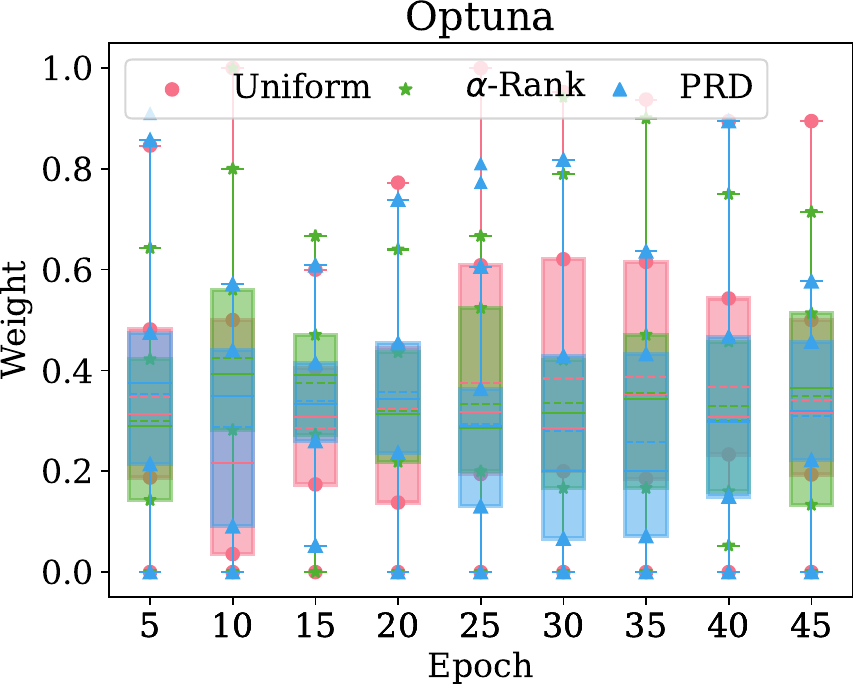}
    \includegraphics[width=0.49\columnwidth]{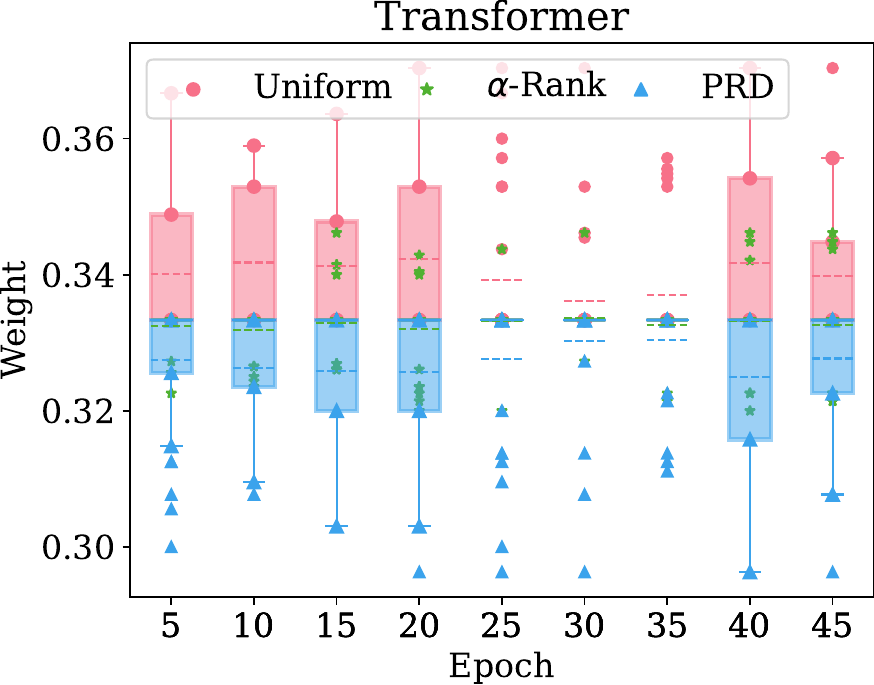}
    \caption{Weights of meta-solvers in \textbf{NFG 150$\times$300}.}
    \label{fig:nfg_150_300}
\end{figure}

\begin{figure}[hbt!]
    \centering
    \includegraphics[width=0.49\columnwidth]{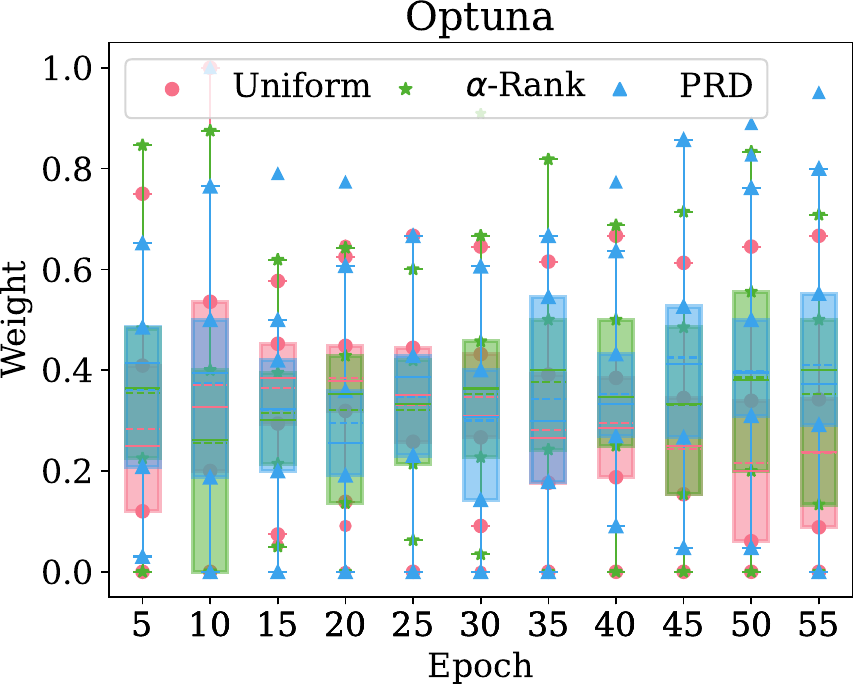}
    \includegraphics[width=0.49\columnwidth]{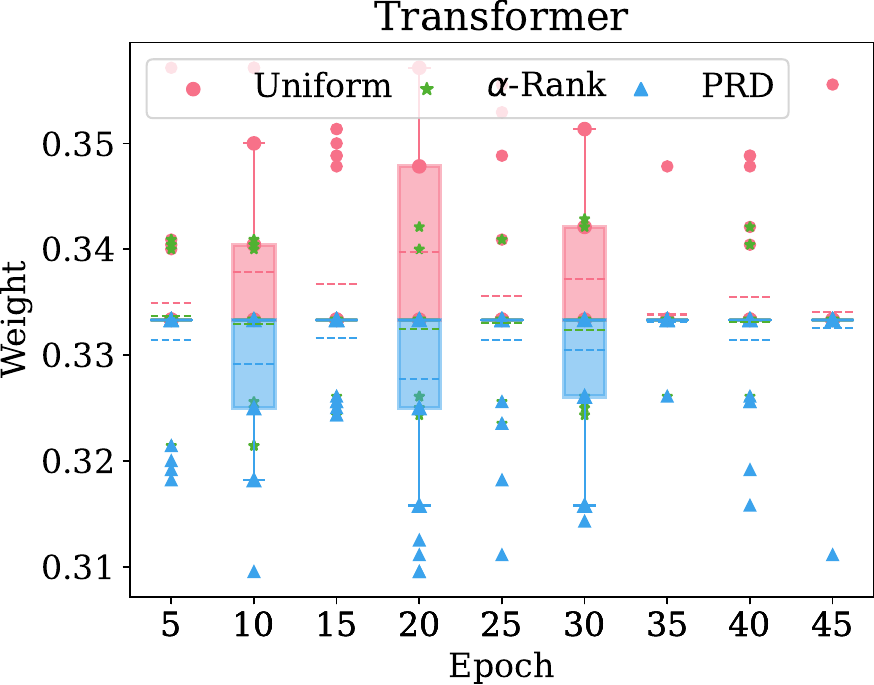}
    \caption{Weights of meta-solvers in \textbf{Leduc Poker}.}
    \label{fig:w_leduc}
\end{figure}

\begin{figure}[hbt!]
    \centering
    \includegraphics[width=0.49\columnwidth]{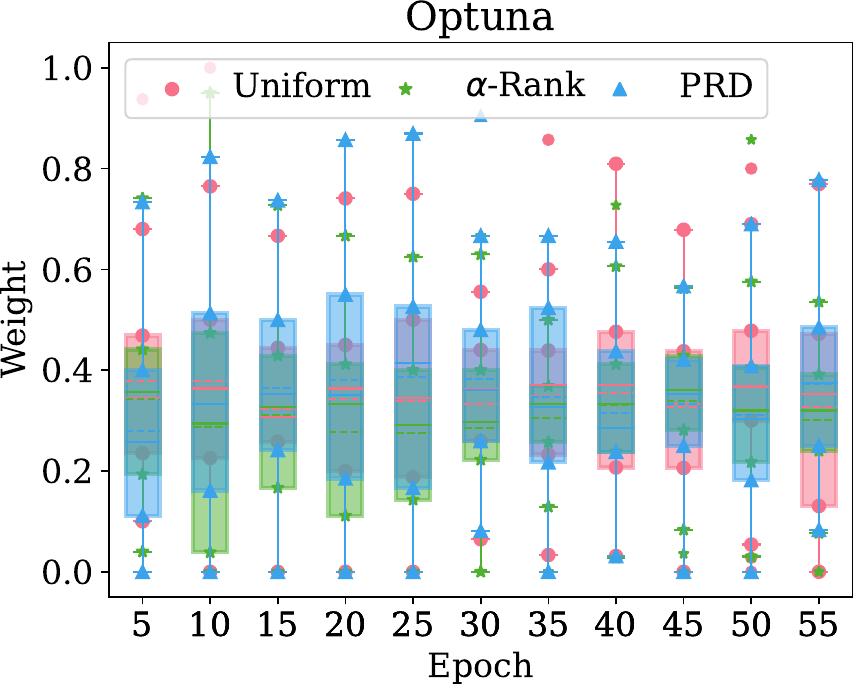}
    \includegraphics[width=0.49\columnwidth]{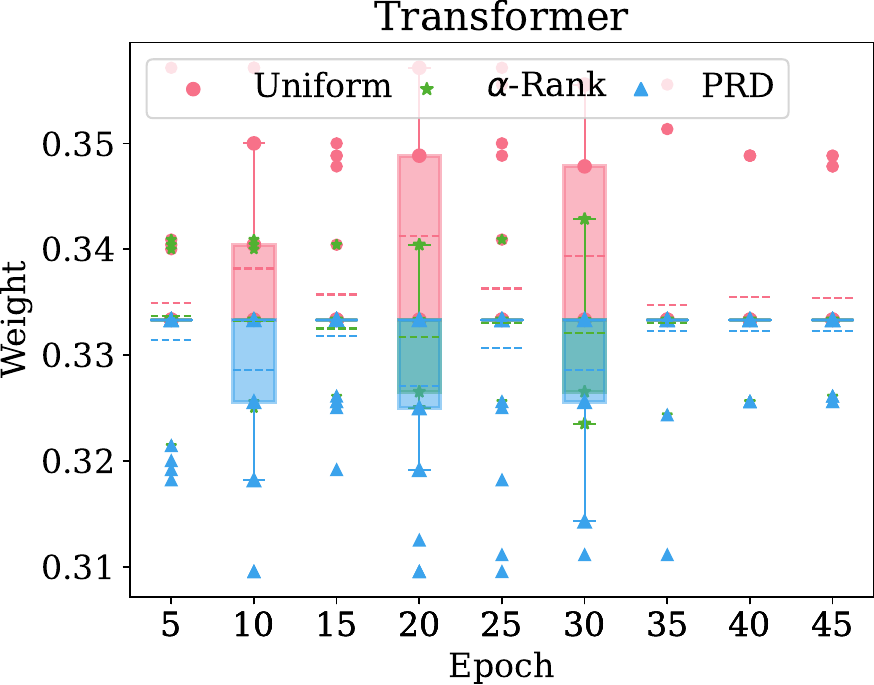}
    \caption{Weights of meta-solvers in \textbf{Goofspiel}.}
    \label{fig:w_goofspiel}
\end{figure}

\begin{figure}[hbt!]
    \centering
    \includegraphics[width=0.49\columnwidth]{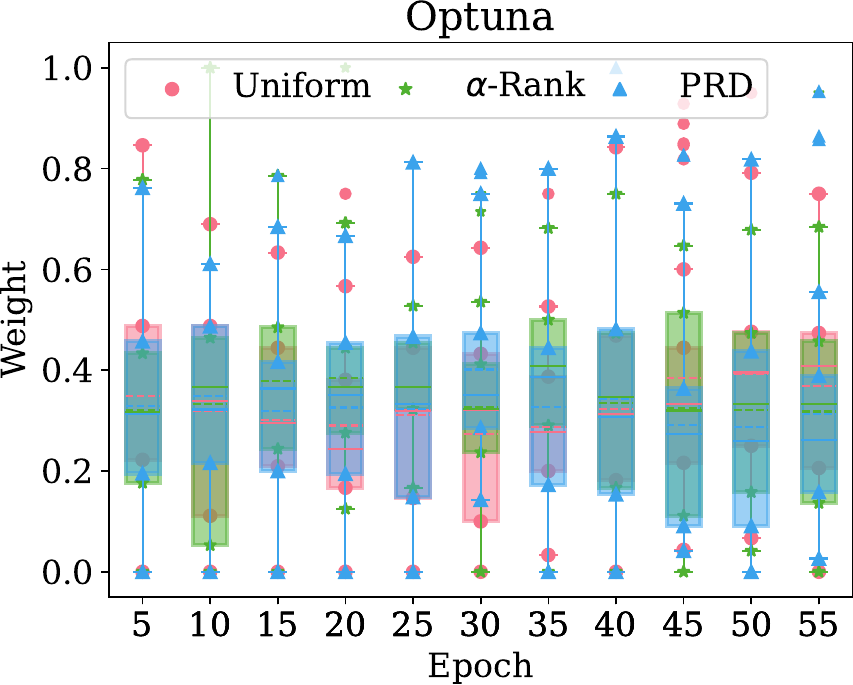}
    \includegraphics[width=0.49\columnwidth]{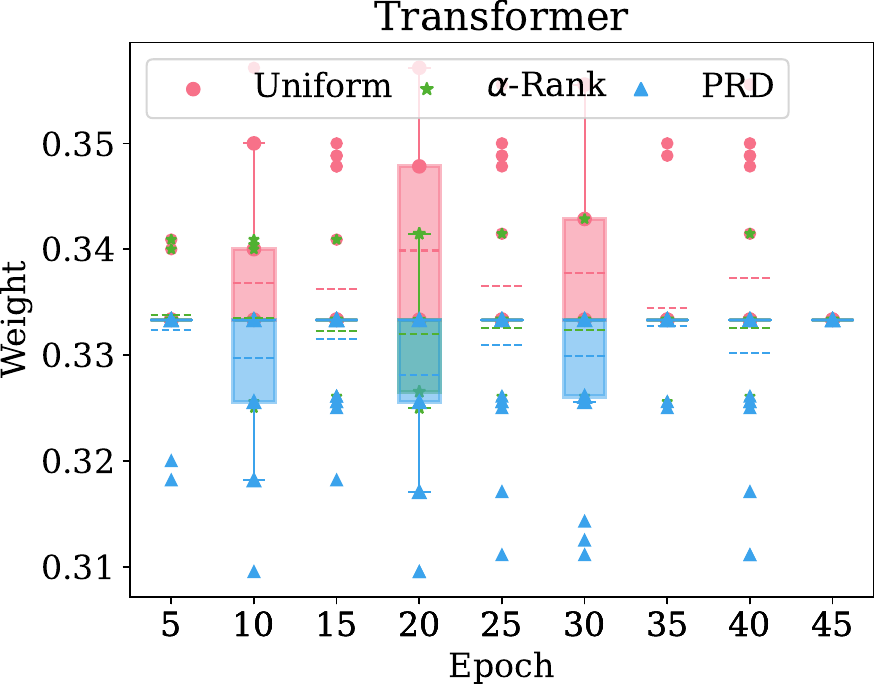}
    \caption{Weights of meta-solvers in \textbf{Liar's Dice}.}
    \label{fig:w_liars_dice}
\end{figure}

\begin{figure}[hbt!]
    \centering
    \includegraphics[width=0.49\columnwidth]{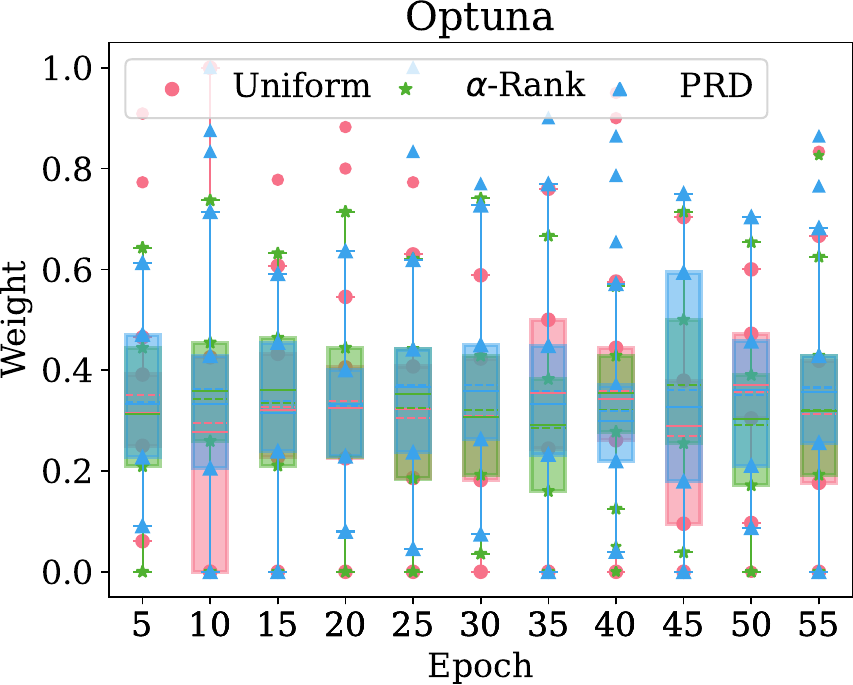}
    \includegraphics[width=0.49\columnwidth]{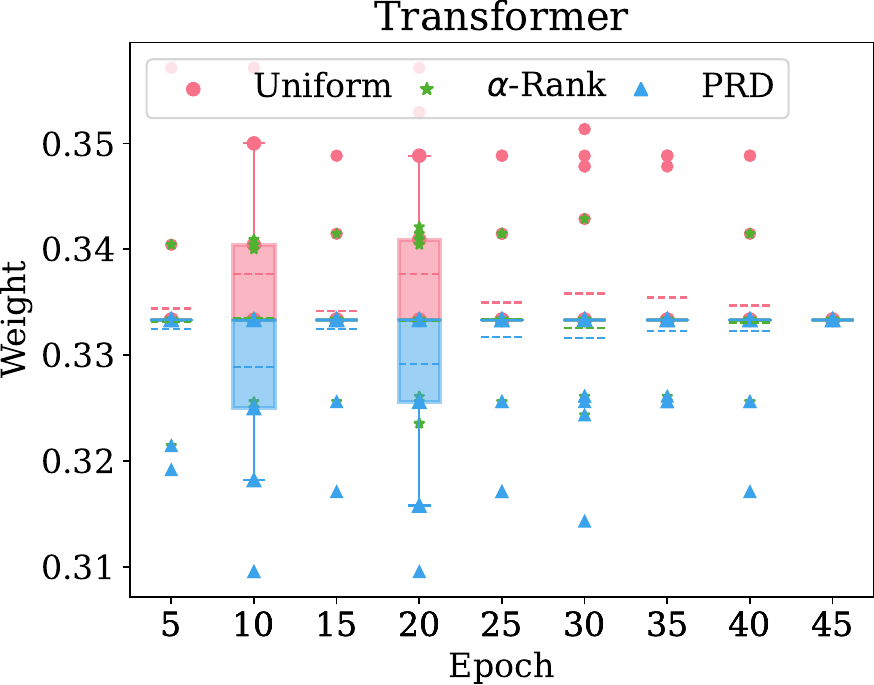}
    \caption{Weights of meta-solvers in \textbf{Negotiation}.}
    \label{fig:w_negotiation}
\end{figure}

\begin{figure}[hbt!]
    \centering
    \includegraphics[width=0.49\columnwidth]{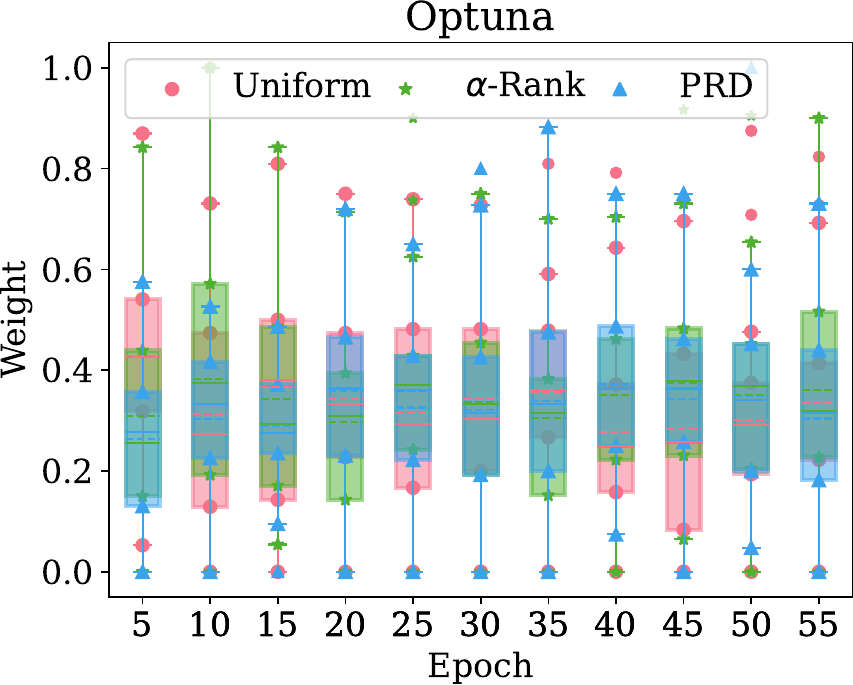}
    \includegraphics[width=0.49\columnwidth]{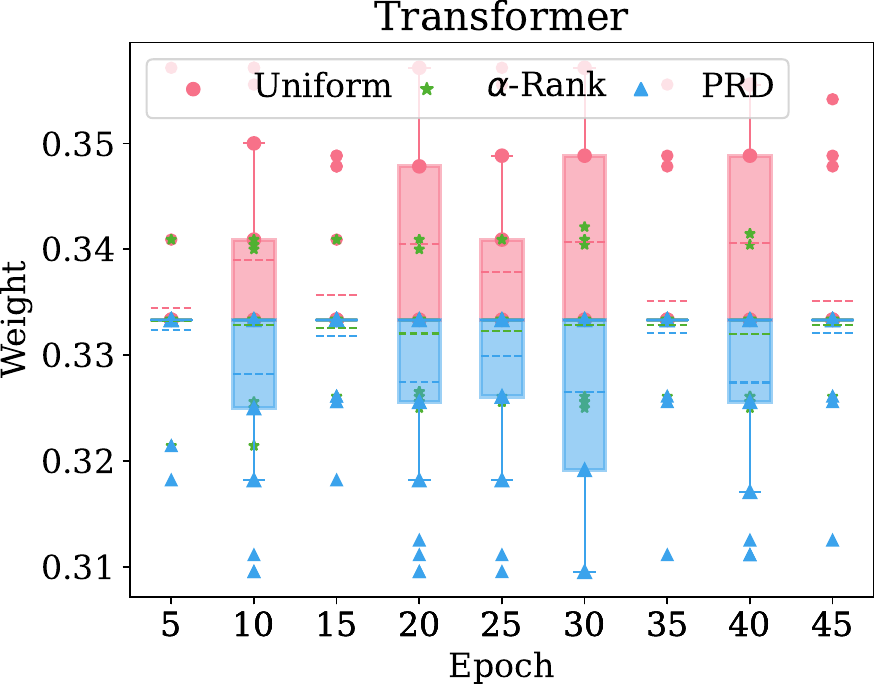}
    \caption{Weights of meta-solvers in \textbf{Tic-Tac-Toe}.}
    \label{fig:w_tic_tac_toe}
\end{figure}

\begin{figure}[hbt!]
\centering
\subfigure[Liar's Dice]{\includegraphics[width=0.49\columnwidth]{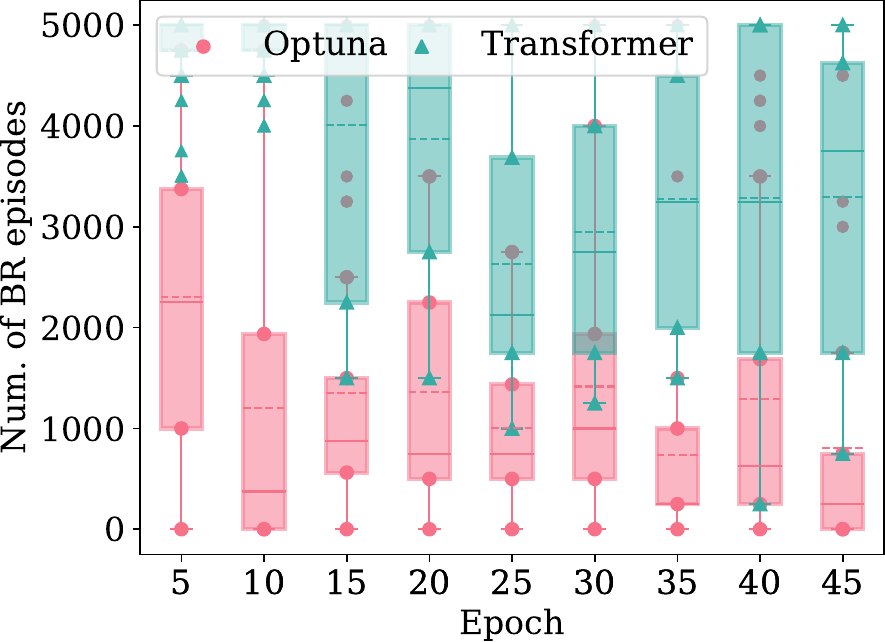}}
\subfigure[Negotiation]{\includegraphics[width=0.49\columnwidth]{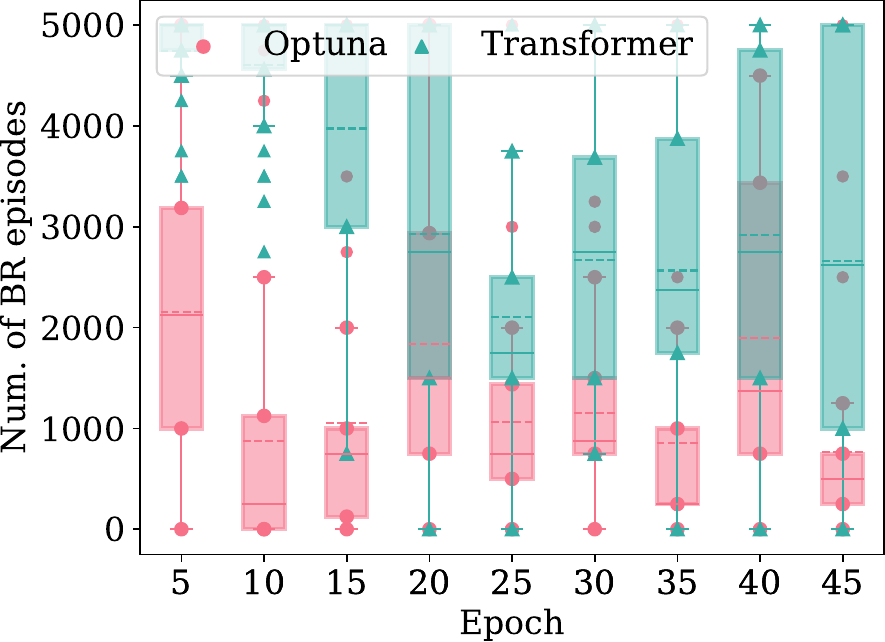}}
\subfigure[Tic-Tac-Toe]{\includegraphics[width=0.23\textwidth]{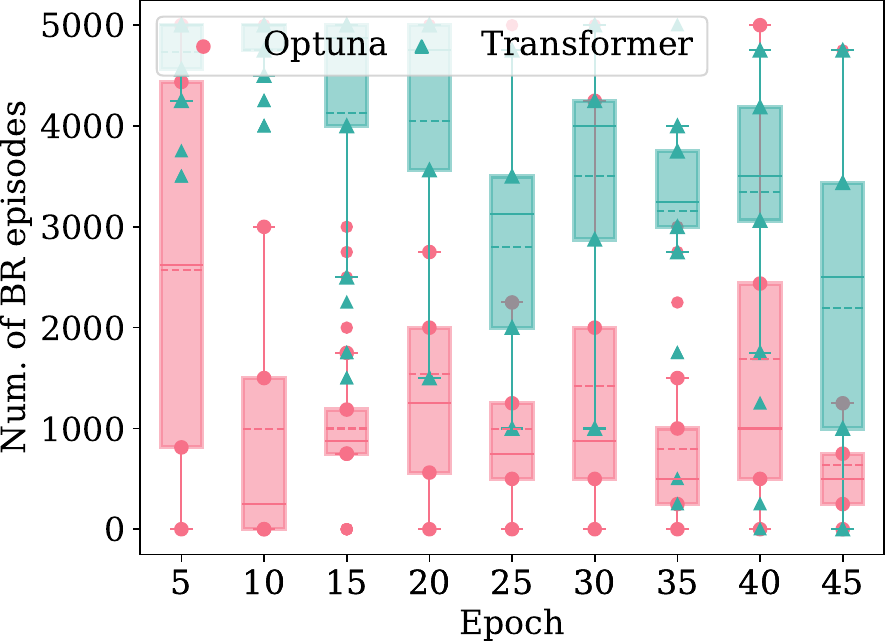}}
\caption{Numbers of BR training episodes determined by Optuna and Transformer in \textbf{Liar's Dice}, \textbf{Negotiation}, and \textbf{Tic-Tac-Toe}. The boxes are obtained by counting 30 independent runs.}
\label{fig:br-time-ttt}
\end{figure}

\noindent\textbf{More Experimental Results.} For NFGs $\{150\times150, 250\times250, 100\times200, 150\times300\}$, the weights of the three meta-solvers (Uniform, PRD, and $\alpha$-Rank) determined by Optuna and Transformer are shown in Figure~\ref{fig:nfg_150_150} to Figure~\ref{fig:nfg_150_300}, respectively. For Leduc Poker, Goofspiel, Liar's Dice, Negotiation, and Tic-Tac-Toe, the weights of the three meta-solvers determined by Optuna and Transformer are shown in Figure~\ref{fig:w_leduc} to Figure~\ref{fig:w_tic_tac_toe}, respectively. For Liar's Dice, Negotiation, and Tic-Tac-Toe, the numbers of BR training episodes determined by
Optuna and Transformer are shown in Figure~\ref{fig:br-time-ttt}.

\clearpage

\end{document}